\pdfoutput=1
\documentclass{article}

\newcommand{\real}{\mathbb{R}}

\newcommand{\bI}{\mathbf{I}}

\newcommand{\model}{PolyTopoBench}

\PassOptionsToPackage{numbers}{natbib}
\usepackage[eandd, final]{neurips_2026}

\usepackage[utf8]{inputenc} 
\usepackage[T1]{fontenc}    
\usepackage{hyperref}       
\usepackage{url}            
\usepackage{booktabs}       
\usepackage{amsfonts}       
\usepackage{nicefrac}       
\usepackage{microtype}      
\usepackage{xcolor}         
\usepackage{amsmath}
\usepackage{amssymb}
\usepackage{mathtools}
\usepackage{amsthm}
\usepackage{graphicx}
\usepackage{enumitem}
\usepackage{multirow}
\usepackage{array}
\theoremstyle{definition}
\newtheorem{definition}{Definition}[]

\usepackage{titlesec}
\titlespacing*{\section}      {0pt}{-2pt}{-2pt}
\titlespacing*{\subsection}   {0pt}{-2pt}{-2pt}
\titlespacing*{\subsubsection}{0pt}{4pt}{2pt}
\titlespacing*{\paragraph}    {0pt}{0pt}{0pt}

\AtBeginDocument{%
  \abovedisplayskip=2pt
  \belowdisplayskip=4pt
  \abovedisplayshortskip=2pt
  \belowdisplayshortskip=2pt
}

\title{
\model{}: A Benchmark 
for Complex Vector Polygon Generation from Remote Sensing Imagery
}

\author{%
  \textbf{Zeping Liu}$^{1}$ \quad
  \textbf{Ni Lao}$^{1}$ \quad
  \textbf{Weiwei Sun}$^{2}$ \quad
  \textbf{Gil Wolff}$^{2}$ \\
  \textbf{Yiqun Xie}$^{3}$ \quad
  \textbf{Liang Zhao}$^{4}$ \quad
  \textbf{Junfeng Jiao}$^{1}$ \quad
  \textbf{Gengchen Mai}$^{1,}$\thanks{Corresponding author.} \\[0.6em]
  \normalfont
  $^{1}$The University of Texas at Austin \quad
  $^{2}$Amazon \quad
  $^{3}$University of Maryland \quad
  $^{4}$Emory University \\[0.4em]
  \normalfont\small
  \texttt{\{zeping.liu, nlao\}@utexas.edu}, \quad
  \texttt{weiwei.sun3@gmail.com}, \quad
  \texttt{wolffg@amazon.com}, \\
  \normalfont\small
  \texttt{xie@umd.edu}, \quad
  \texttt{liang.zhao@emory.edu}, \quad
  \texttt{\{jjiao, gengchen.mai\}@austin.utexas.edu} \\
}

\begin{document}

\maketitle

\begin{abstract}

\textbf{Vector polygon generation} converts visual inputs, e.g., remote sensing (RS) images, into vectorized polygonal geometries, supporting applications such as autonomous driving, vector map construction, and remote sensing. Early pipelines predict raster masks and post-process them into polygons, which prevents end-to-end optimization and may miss small objects or introduce inaccurate vertices. Recent methods directly generate vector polygons, but most focus on simple exterior contours, while they either cannot represent complex polygons with holes or fail to preserve their topology. In this paper, we propose \textbf{\model{}}, a unified evaluation framework for vector polygon generation from RS images with explicit emphasis on complex polygons. \model{} evaluates both exterior and interior rings, and benchmarks 11 representative methods, including segmentation-based polygonization pipelines, vision foundation model baselines, and specialized vector polygon generators, on two RS-image datasets covering buildings, roads, vegetation, and unvegetated regions. Experiments show that existing methods often recover simple exterior boundaries but degrade substantially on polygons with holes or multiple rings. These results reveal complex polygon generation as an unresolved challenge and motivate topology-aware benchmarks and model designs. Code and data are available at \url{https://github.com/seai-lab/PolyTopoBench}.



\end{abstract}

\section{Introduction}  \label{sec:intro}
\textbf{Vector polygon generation from remote sensing images} converts visual observations into editable geometric vectors, and has many applications such as urban planning~\cite{liu2023china, li2026satellite}, environmental monitoring~\cite{zhu2022remote, ao2024national}, and disaster response~\cite{najafi2024high,xu2022seismic}.
A useful polygon prediction must therefore capture not only the visible extent of an object, but also the structure that makes the geometry valid and interpretable. 
In this paper, we distinguish between \emph{simple polygons}, comprising a single exterior ring, and \emph{complex polygons}, each of which consists of one exterior ring and one or more interior rings (holes). 
This distinction is especially important because complex polygons encode essential topological and semantic information that cannot be recovered merely from the exterior boundary alone. Interior rings indicate regions that are explicitly excluded from the current object, and their presence often corresponds to meaningful real-world spatial structure rather than noise. 
For example, buildings may contain courtyards or enclosed voids, roads may encircle traffic islands or land parcels, and land parcels may contain holes induced by other classes. 
Missing these hole structures fundamentally misrepresents the spatial extent and function of the object, and can introduce systematic errors in downstream tasks such as area and shape descriptor computation \cite{yan2019graph}, spatial relation prediction \cite{mai2023towards,ji2025foundation}, and geographic question answering \cite{yu2025spatial,bao2026spatial,mai2018poireviewqa}. 

Despite the importance of this multi-ring representation, most existing polygon-generation methods~\cite{girard2021polygonal,zorzi2022polyworld,jiao2026acpv} and evaluation benchmarks~\cite{maggiori2017can,ji2018fully, sulzer2025p} still \textbf{treat polygonization mainly as an exterior-boundary problem}. Classical pipelines predict raster masks and then simplify their contours into polygons \cite{zhao2018building, zorzi2021machine} or use simplified bounding boxes \cite{xie2020locally}, which often results in heavy artifacts, such as irregular and inaccurate boundaries due to the lack of explicit geometric modeling. To address these, recent specialized polygon-generation methods learn vertex~\cite{xu2023hisup,jiao2026acpv,jiao2025roipoly}, contour~\cite{zhang2025global,wang2025holitracer}, frame-field~\cite{girard2021polygonal}, graph~\cite{zorzi2022polyworld}, or token-sequence~\cite{adimoolam2025pix2poly} representations to directly generate vector outputs. These models achieve strong results on common benchmarks, especially for building-footprint extraction. However, these standard evaluation settings remain limited in three ways. 
The targeted geospatial objects are often dominated by buildings, whose instances are comparatively compact and frequently simple. 
Many benchmarks~\cite{maggiori2017can,ji2018fully, luo2023diverse, sulzer2025p} contain few complex polygons with interior rings, making it difficult to evaluate whether methods can handle multi-ring topology. 
Furthermore, many existing evaluations~\cite{lin2014microsoft,cheng2021boundary,avbelj2014metric,girard2021polygonal} often reduce polygon geometry to a single closed contour, without explicitly distinguishing exterior from interior rings, or simply omitting interior rings altogether. 
As a result, a method can appear successful under common scores while producing geometries that are incomplete or topologically incorrect for downstream geospatial applications.

These limitations call for benchmarks that explicitly evaluates  models' capabilities to generate complete multi-ring vector geometries, rather than simple exterior contours. To this end, we introduce \textbf{\model{}}, a benchmark and evaluation framework for complex vector polygon generation from remote-sensing imagery. To the best of our knowledge, \model{} is the first benchmark in remote-sensing polygon generation that places complex polygons with interior rings, rather than only simple polygons or exterior boundaries, at the center of evaluation. It goes beyond the common building and crop field-only setting~\cite{maggiori2017can,ji2018fully,sulzer2025p,kerner2025fields} by covering four object categories: buildings, roads, vegetation, and unvegetated regions. Based on Inria buildings~\cite{maggiori2017can} and Deventer land-cover polygons~\cite{jiao2026acpv}, we incorporate OpenStreetMap~\cite{OpenStreetMap} annotations and manually correct them to align with the imagery and recover missing interior rings. The resulting benchmark contains nearly 300K polygon instances, more than 10K complex polygons, and over 42K interior rings.

\model{} further includes 11 diverse and publicly available vector polygon generation models, spanning segmentation-then-polygonization pipelines~\cite{ronneberger2015u,he2017mask,zorzi2021machine}, vision foundation model-based polygonization~\cite{muhawenayo2026prue}, representation-based polygon models~\cite{xu2023hisup,jiao2026acpv,girard2021polygonal,zhang2025global,wang2025holitracer}, and direct polygon generation methods~\cite{adimoolam2025pix2poly,zorzi2022polyworld,jiao2025roipoly}. Finally, \model{} introduces hole-aware evaluation metrics that separately assess overall polygon geometries, exterior rings, and interior rings for more robust topology evaluation.

In summary, our contributions are as follows:
\vspace{-0.2cm}
\begin{itemize}[leftmargin=*,itemsep=1pt, parsep=0pt, topsep=0pt]
    \item We introduce \model{}, a unified remote-sensing benchmark that places \emph{complex} vector polygon generation, rather than simple exterior-boundary extraction.
    \item We construct a large-scale and multi-category benchmark with nearly 300K polygon instances, over 10K complex polygons, more than 42K interior rings, and four classes of geospatial objects.
    \item We provide a comprehensive and reproducible comparison of 11 diverse, publicly available vector polygon generation models, and propose a hole-aware evaluation protocol that separately assesses exterior and interior rings through ring-aware boundary metrics.
    \item We reveal a shared limitation of current polygon-generation methods -- most existing evaluation metrics for polygon generation models focus mainly on region overlap or exterior-boundary quality, which are insufficient since they do not guarantee correct interior rings; and no existing method reliably solves complex polygon generation problem based on our evaluation.
\end{itemize}

\section{Related Work}
\label{sec:related}

\subsection{Vector Polygon Generation}
\label{sec:related_methods}

Vector polygon generation has been studied broadly in computer vision, evolving from raster post-processing to structured vector prediction. 
Early pipelines typically follow a segmentation-then-polygonization paradigm, where semantic segmentation~\cite{tajbakhsh2016convolutional} or instance segmentation models~\cite{he2017mask} first produce raster masks, and post-processing algorithms then extract, simplify, or regularize contours into vector polygons~\cite{zhao2018building,zorzi2021machine}. 
Beyond mask-based pipelines, later methods model object boundaries more explicitly, including contour-deformation methods that deform initialized contours toward object boundaries~\cite{kass1988snakes,peng2020deep,liang2020polytransform,liu2021dance}, RNN-based methods that sequentially predict ordered vertices~\cite{castrejon2017annotating,acuna2018efficient}, graph-based methods that represent object boundaries as graphs and iteratively refine polygon vertices and edges~\cite{ling2019fast}. These general-purpose vision methods provide important foundations for editable vector output, but they are not specifically optimized for geospatial objects, whose boundaries can be long, irregular, multi-scale, and topologically complex \cite{zheng2020foreground,garnot2021panoptic}.

Recent remote-sensing methods introduce domain-specific geometric representations for polygon extraction. Frame-field methods predict local orientation fields to guide polygonization and improve boundary regularity~\cite{girard2021polygonal}. Graph-based methods represent vector structures as connected vertices and edges~\cite{zorzi2022polyworld}, and vertex- or representative-point-based methods detect corners, control points, or boundary keypoints before assembling polygons~\cite{huang2021oec,liu2022building,xu2023hisup,hu2023polybuilding, jiao2025roipoly,jiao2026acpv}. Recently, LLM-based methods have also been explored for vector polygon generation~\cite{zhang2026vectorllm}, showing promising performance. These representations have improved geometric accuracy and regularity for remote-sensing objects, but they still provide limited support for complex polygons with interior rings, and often fail to explicitly model multi-ring topology.

\subsection{Vector Polygon Generation Benchmarks and Evaluation}
\label{sec:related_benchmarks}

Existing geospatial polygon benchmarks are mainly built around buildings and crop fields. Representative building-footprint datasets include Inria~\cite{maggiori2017can}, CrowdAI~\cite{mohanty2020deep}, WHU~\cite{ji2018fully}, and SpaceNet~\cite{van2018spacenet}, while crop-field polygon datasets include AI4SmallFarms~\cite{persello2023ai4smallfarms}, Fields of The World (FTW)~\cite{kerner2025fields}, and AI4Boundaries~\cite{d2023ai4boundaries}. Other category-specific datasets have also been introduced for water bodies, such as GLH-Water~\cite{li2024glh}, and roads, such as VHR-Road~\cite{wang2025holitracer}. More recently, Deventer-512~\cite{jiao2026acpv} extends polygon vectorization to multiple land-cover categories, including buildings, roads, vegetation, water, and unvegetated areas.

Despite these efforts, existing benchmarks provide limited support for evaluating complex polygon generation. Many datasets focus on category-level region extraction rather than multi-ring vector geometry topology, and common metrics such as IoU, AP, BIoU, POLIS, and MTA~\cite{lin2014microsoft,cheng2021boundary,avbelj2014metric,girard2021polygonal,muhawenayo2026prue} mainly measure region overlap or aggregated boundary quality. Since they usually do not explicitly distinguish exterior rings from interior rings, methods can achieve strong benchmark scores while still wrongly representing complex polygons.

\section{\model{}}  \label{sec:method}

\begin{figure*}[t]
    \centering
    \includegraphics[width=\textwidth]{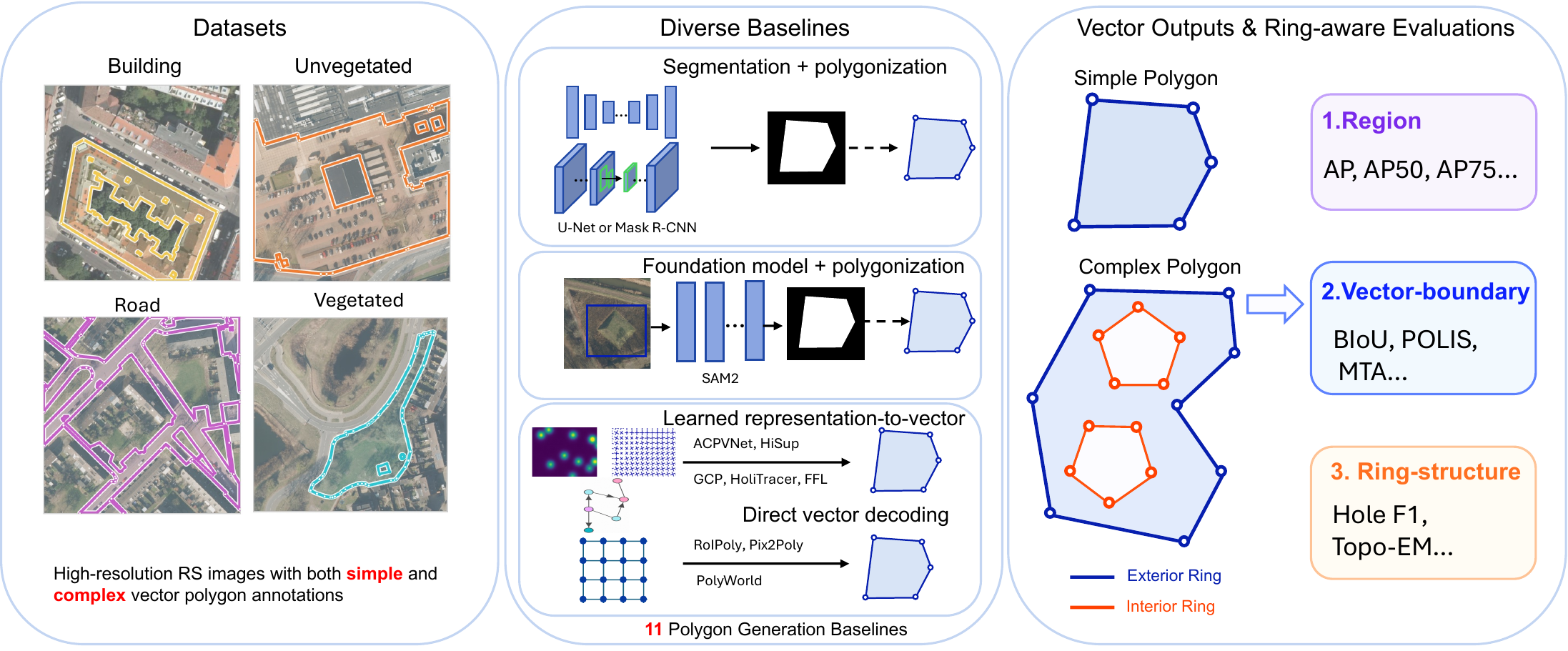}
    \caption{
    \textbf{Overview of \model{}.}
    \model{} evaluates complex vector polygon generation from remote sensing imagery.
    It consists of diverse remote sensing imagery datasets with simple and complex polygon annotations on different geographic features, including \emph{buildings, roads, unvegetated regions, and vegetated regions}. It supports eleven diverse polygon generation baselines.
    A unified evaluator is developed,
    which jointly considers \emph{polygonal region agreement, vector-boundary quality, and ring-structure correctness}. 
    }
    \label{fig:framework}

\end{figure*}

\subsection{Task and Polygon Representation}
\label{sec:method_task}

\model{} aims to systematically evaluate diverse vector polygon generation models from remote sensing imagery. Unlike most existing polygon generation methods, which focus on simple polygon generation, we systematically evaluate 11 polygon generation models on both simple polygon generation and complex polygon generation (i.e., polygons with one or multiple holes). A unified evaluator is developed to evaluate and compare diverse methods, where novel evaluation metrics have been proposed to evaluate polygon generations from three distinct perspectives: region agreement, vector-boundary quality, and ring-structure correctness. 
Figure~\ref{fig:framework} summarizes the benchmark design. Given an input image $\bI \in \real^{H \times W \times 3}$, each method is converted to a set of scored polygon instances
\begin{equation}
    \widehat{\mathcal{P}}(\bI)=\{(\hat{c}_i,\hat{s}_i,\hat{G}_i)\}_{i=1}^{N},
\end{equation}
where $\hat{c}_i$ is the predicted category, $\hat{s}_i$ is a confidence score, and $\hat{G}_i$ is a single polygonal instance geometry with exactly one exterior ring and zero or more interior rings. The corresponding ground-truth set is
\begin{equation}
    \mathcal{P}(\bI)=\{(c_j,G_j)\}_{j=1}^{M}.
\end{equation}
Here, $c_j$ and $G_j$ denote the ground-truth category and polygonal geometry. 

For a polygon instance $G$, we denote its exterior ring by $e(G)$ and its set of interior rings by $\mathcal{I}(G)$. Let
\begin{equation}
    H(G)=|\mathcal{I}(G)|, \qquad R(G)=1+H(G),
\end{equation}
where $H(G)$ is the number of holes and $R(G)$ is the total number of rings. We call $G$ \emph{simple} when $H(G)=0$ and \emph{complex} when $H(G)\geq 1$. Thus, unlike raster segmentation benchmarks that mainly evaluate the occupied image region, \model{} evaluates the full vector polygon structure. Both exterior and interior rings are part of the target geometry, so a prediction with a correct outer boundary but missing holes is still treated as an incomplete polygon.

\subsection{Datasets and Benchmark Tasks}
\label{sec:method_data}

\model{} uses two aerial-image datasets that have different geometric objects with diverse classes. The Inria building dataset \cite{maggiori2017can} evaluates building-footprint generation. Buildings are often compact and piecewise regular, but courtyards, inner voids, and clipped structures create multi-ring instances that cannot be represented by a simple polygon with one exterior boundary alone. Since the original Inria release provides binary masks, we align OpenStreetMap \cite{OpenStreetMap} vector annotations to the image tiles, correct them manually, and use the resulting polygons as vector polygon ground truth, which contains both simple and complex polygonal geometries. The Deventer \cite{jiao2026acpv} data provide multi-class land-cover vector polygons on $512\times512$ aerial patches. In this paper, we focus on the road, vegetation, and unvegetated classes because they contain richer complex-polygon structures.

Table~\ref{tab:dataset_complexity} summarizes the statistics of hole-bearing polygons that appear in all four tasks. These statistics motivate a systematic evaluation that separates polygonal geometry exterior-boundary accuracy from interior-ring recovery.

\begin{table*}[t]
\centering
\small
\caption{
Dataset complexity statistics for the four single-class \model{} tasks. \emph{Instances} denotes the number of total polygon annotations in each dataset/class. \emph{Simple} and \emph{Complex} denote instances with $H(G)=0$ and $H(G)\geq 1$. \emph{Holes} counts all interior rings in complex instances. \emph{Holes / complex} is the average number of holes per complex instance. \emph{Avg. verts} is the average number of vertices per instance, including exterior and interior rings and excluding the repeated closing point.
}
\label{tab:dataset_complexity}
\setlength{\tabcolsep}{4.2pt}
\begin{tabular}{llrrrrrr}
\toprule
Dataset & Class & Instances & Simple & Complex & Holes & Holes / complex & Avg. verts \\
\midrule
Inria & Building & 254,133 & 248,621 & 5,512 & 19,854 & 3.60 & 10.27 \\
Deventer & Road & 4,533 & 3,467 & 1,066 & 9,992 & 9.37 & 62.57 \\
Deventer & Vegetation & 23,517 & 22,658 & 859 & 1,477 & 1.72 & 11.98 \\
Deventer & Unvegetated & 16,552 & 13,610 & 2,942 & 10,749 & 3.65 & 21.53 \\
\bottomrule
\end{tabular}
\end{table*}

\begin{table*}[t]
\centering
\scriptsize
\caption{
Comparing polygon generation baselines. 
\emph{Ring mode} is explicit (denoted as \emph{Exp.})
when holes are natively modeled and implicit (denoted as \emph{Imp.}) otherwise.
}
\label{tab:baseline_taxonomy}
\setlength{\tabcolsep}{3.2pt}
\begin{tabular}{@{}p{0.2\textwidth}>{\centering\arraybackslash}
p{0.67\textwidth}p{0.055\textwidth}@{}}
\toprule
Method & Type and Description & Ring \\
\midrule
& \textbf{Seg: Segmentation-then-polygonization}& \\ 
U-Net + Poly.~\cite{ronneberger2015u,zorzi2021machine}  
& Predict a semantic mask and convert its contours into polygons with rule-based simplification. & Imp. \\
Mask R-CNN + Poly.~\cite{he2017mask}   
& Predict instance masks and convert each mask contour into a simplified polygon. & Imp. \\ \hline
& \textbf{FM: Foundation-model-assisted polygonization}& \\ 
SAM2 + Poly.~\cite{muhawenayo2026prue}                  
& Generate masks with SAM2 prompts or priors and polygonize the resulting mask contours. & Imp. \\ \hline
& \textbf{Rep.: Learned representation-to-vector}& \\ 
HiSup~\cite{xu2023hisup}                                
& Learn masks, vertices, and attraction fields, then extract mask contours to vertices. & Imp. \\
ACPV-Net~\cite{jiao2026acpv}                            
& Learn semantic masks and vertex heatmaps, then reconstruct a shared-boundary planar graph. & Exp. \\
FFL~\cite{girard2021polygonal}                          
& Learn masks and frame fields, then optimize contours along the learned directions. & Exp. \\
GCP~\cite{zhang2025global}                              
& Refine mask contours with a transformer and simplify them with global collinearity. & Exp. \\
HoliTracer~\cite{wang2025holitracer}                    
& Segment large images, reform mask contours, and trace vertices with a polygon sequence model. & Imp. \\ \hline
& \textbf{Direct: Direct vector decoding}& \\ 
Pix2Poly~\cite{adimoolam2025pix2poly}                   
& Decode polygon vertices as a transformer token sequence. & Imp. \\
PolyWorld~\cite{zorzi2022polyworld}                     
& Predict vertices and connect them with a graph model to form polygons. & Imp. \\
RoIPoly~\cite{jiao2025roipoly}                          
& Decode ordered vertices from each proposal region using vertex and logit queries. & Imp. \\
\bottomrule
\end{tabular}
\end{table*}

\subsection{Polygon Ring-Aware Evaluation Metrics}
\label{sec:method_metrics}

Prior polygon benchmarks treat a model prediction either as an occupied image region with evaluation metrics such as IoU and  AP \cite{lin2014microsoft,van2018spacenet} or as a single polygon boundary contour with evaluation metrics such as BIoU, POLIS, and MTA \cite{avbelj2014metric,liu2022building}. Both scores conflate errors on the exterior ring with errors on interior rings, and neither records whether the predicted polygon has the correct hole count or a valid topology. A method that returns only the exterior of a hole-bearing polygon can therefore obtain a near-perfect AP50. However, polygon holes sometimes carry an important meaning (e.g., farmland ownership), and failing to capture these holes can lead to significant issues in downstream applications \cite{mai2023towards,yu2025spatial}. 

Our \model{} closes this gap by evaluating a polygon as a set of role-specific rings rather than as an image region or a single contour. We organize metrics into three complementary groups and highlight our novel contributions in each: \textbf{region agreement} metrics reuse standard AP50 for comparability; \textbf{vector-boundary quality} metrics keep BIoU, POLIS, and MTA but report them under three novel \emph{ring-aware scopes} -- Overall, Exterior,  and Hole -- that localize error to the exterior ring or the holes; and \textbf{ring-structure correctness} metrics include two novel topology-oriented metrics, \emph{Hole-F1} and \emph{Topo-EM}, built on a \emph{role-specific ring matching} protocol. Full details are in Appendix~\ref{app:metric_details}; below we summarize metrics and focus on novel metrics proposed in \model{}.

\textbf{Region agreement metrics. }
For comparability with prior polygon-generation benchmarks, we report \emph{AP50}. AP50 measures instance-level area overlap and is ring-agnostic by construction: a prediction that recovers the exterior boundary but misses every hole still receives a near-perfect score, which motivates the ring-localized metrics below.

\textbf{Vector-boundary quality metrics.}
We report three standard contour metrics---\emph{BIoU}~\cite{cheng2021boundary},
\emph{POLIS}~\cite{avbelj2014metric}, and \emph{MTA}~\cite{girard2021polygonal}%
---which measure boundary-buffer overlap, vertex-to-boundary distance, and
tangent-angle deviation, respectively (BIoU is higher-is-better, while POLIS and MTA are lower-is-better). Our contribution is to evaluate each of them under three \emph{ring-aware scopes} that restrict which rings of a matched instance pair enter the computation: \emph{Overall} (O) uses all rings of the matched pair (exterior and interior), \emph{Ext.} (E) uses only the paired exterior rings, and \emph{Hole} (H) uses interior-ring pairs produced by role-specific matching (below). The three scopes split each boundary score into contributions from the outer boundary versus recovered holes, in which a single polygon-level number silently merges.

\textbf{Ring-structure correctness metrics.}
Ring-structure metrics evaluate whether interior holes and the complete polygon topology are recovered. They require a correspondence between predicted and ground-truth rings, which we obtain by \emph{role-specific ring matching}: for an IoU-matched instance pair $(\hat{G},G)$, the two exterior rings are paired directly, while interior rings in $\mathcal{I}(\hat{G})$ and $\mathcal{I}(G)$ are matched one-to-one via Hungarian assignment with per-ring BIoU as the cost, accepting a hole pair only when its BIoU exceeds a fixed threshold. This prevents a predicted hole from being counted against a ground-truth hole in a different part of the polygon. Please see Definition \ref{def:role_ring_math} in Appendix \ref{app:metric_details} for detailed descriptions. Based on this, we introduce two metrics:

\begin{definition}[Hole-F1]
\emph{Hole-F1} treats each interior ring as a detection object. Let $\mathrm{TP}_{\mathrm{h}}$ be the number of accepted hole matches, $\mathrm{FP}_{\mathrm{h}}$ the number of unmatched predicted holes, 
and $\mathrm{FN}_{\mathrm{h}}$ the number of unmatched ground-truth holes. Hole-F1 is defined as
\begin{equation}
    \operatorname{Hole\mbox{-}F1}
    =
    \frac{2\mathrm{TP}_{\mathrm{h}}}
         {2\mathrm{TP}_{\mathrm{h}}+\mathrm{FP}_{\mathrm{h}}+\mathrm{FN}_{\mathrm{h}}}.
\end{equation}
Unlike IoU or exterior BIoU, Hole-F1 collapses to zero whenever a method never emits interior rings, directly exposing the single-exterior-contour failure mode.
\label{def:hole_f1}
\end{definition}

\begin{definition}[Topo-EM]
\emph{Topo-EM} (topology exact match) is a stricter instance-level check on the
entire ring structure. For a matched pair $(\hat{G},G)$, we set $T(\hat{G},G)=1$
only if all the following requirements are met: 1) $\hat{G}$ is a valid polygon; 2) its exterior ring is matched to $e(G)$; 3) the predicted and ground-truth hole counts are equal; and 4) every ground-truth hole is matched one-to-one with a predicted hole. Otherwise, $T(\hat{G},G)=0$.
Let $\mathcal{M}_{\mathrm{inst}}$ be the matched instance pairs and
$\mathcal{U}_{\mathrm{gt}}$, $\mathcal{U}_{\mathrm{pred}}$ the unmatched
ground-truth and predicted instances. Then
\begin{equation}
    \operatorname{Topo\mbox{-}EM}
    =
    \frac{1}{K}
    \sum_{(\hat{G},G)\in\mathcal{M}_{\mathrm{inst}}}
    T(\hat{G},G),
    \qquad
    K=
    |\mathcal{M}_{\mathrm{inst}}|
    +|\mathcal{U}_{\mathrm{gt}}|
    +|\mathcal{U}_{\mathrm{pred}}|.
\end{equation}
\label{def:topo_em}
\end{definition}
\vspace{-0.6cm}
Topo-EM simultaneously punishes missing holes, hallucinated holes, invalid geometry, and mismatched exterior rings. It is therefore used as a strict topology exactness diagnostic rather than as the only leaderboard criterion. In our default evaluation, BIoU uses a one-pixel boundary buffer and ring matches are accepted at a threshold $0.1$. Appendix~\ref{app:metric_sensitivity} studies the sensitivity of Hole-F1 and Topo-EM to stricter ring matching and hole-size filtering.

\subsection{Baselines and Evaluation Protocol}
\label{sec:method_protocol}

As summarized in Table~\ref{tab:baseline_taxonomy}, \model{} evaluates eleven baselines and groups them into four types: 
(1) \emph{Seg.} methods are segmentation-then-polygonization pipelines, including U-Net \cite{ronneberger2015u} and Mask R-CNN \cite{he2017mask}, followed by a deterministic raster-to-vector polygonization \cite{zorzi2021machine}. (2) \emph{FM} methods are vision-foundation-model-assisted polygonization pipelines. We prompt SAM2 with Mask R-CNN bounding boxes and polygonize its masks~\cite{zhao2018building,muhawenayo2026prue}. We also include specialized vector generation models, which are split into \emph{Rep.} and \emph{Direct}. (3) \emph{Rep.} methods include HiSup~\cite{xu2023hisup}, ACPV-Net~\cite{jiao2026acpv}, Frame Field Learning (FFL)~\cite{girard2021polygonal}, GCP~\cite{zhang2025global}, and HoliTracer~\cite{wang2025holitracer}. Each method first predicts an intermediate representation and then decodes it into vector polygons using a representation-specific step (vertex attraction in HiSup, planar-graph assembly in ACPV-Net, frame-field optimization in FFL, transformer contour refinement in GCP, or polygon-sequence tracing in HoliTracer).  (4) \emph{Direct} methods include Pix2Poly~\cite{adimoolam2025pix2poly}, PolyWorld~\cite{zorzi2022polyworld}, and RoIPoly~\cite{jiao2025roipoly}. These methods decode vector primitives or polygon connectivity more directly at the instance level.

Among all baselines, we further distinguish whether interior rings are modeled implicitly or explicitly. \emph{Implicit} hole-aware methods do not natively predict holes as separate polygon rings. Instead, holes are recovered indirectly from masks~\cite{xu2023hisup}, contours~\cite{wang2025holitracer},
ring containment~\cite{adimoolam2025pix2poly,zorzi2022polyworld}, or post-processing~\cite{zorzi2021machine}. \emph{Explicit} hole-aware methods natively model interior rings or shared boundary structures, so holes can be represented as part of the predicted vector topology. In our taxonomy, ACPV-Net, FFL, and GCP are \emph{explicit} hole-aware methods, while the others are \emph{implicit}.

These baselines are intentionally heterogeneous. To compare them fairly, \model{} normalizes every prediction to the same instance-level geometry before evaluation, i.e., a scored polygon with one exterior ring and zero or more interior rings. Native confidence scores are used when available. Score-free methods are assigned a constant score and are primarily compared using non-ranked boundary, ring, and topology metrics.

For \emph{Seg.} methods, we convert raster masks to polygons using contour extraction and Douglas--Peucker simplification. For \emph{FM} methods, SAM2 generates raster masks using Mask R-CNN bbox prior, followed by the same polygonization procedure as \emph{Seg.}. We ablate SAM2 priors under different settings (see Section~\ref{sec:why_fail}). In the \emph{Direct} method group, for methods that output independent closed rings~\cite{adimoolam2025pix2poly,jiao2025roipoly}, we infer polygon structure by ring containment: outer rings become exterior boundaries, while enclosed rings are assigned as interior rings. \model{} does not add holes, repair missing rings, or complete invalid topology before evaluation, so missing interior rings are penalized by Hole-F1 and Topo-EM. Appendix~\ref{app:baseline_impl} provides the detailed settings for each baseline.

\section{Experiment}  \label{sec:exp}

\subsection{Main Benchmark Results}
\label{sec:exp_main_results}

Table~\ref{tab:main_benchmark} reports the full \model{} leaderboard across the four single-class tasks. We discuss the Inria Building task and the three Deventer tasks in turn.

\textbf{Results on the Inria Building dataset.}
The two \emph{Seg.} methods obtain the highest region agreement, with Mask R-CNN + Poly. at AP50 $=0.720$ and U-Net + Poly. at $0.670$, above every \emph{Rep.} and \emph{Direct} baseline. Among \emph{Rep.} methods, GCP ($0.631$), ACPV-Net ($0.575$), and HiSup ($0.496$) are closest, while the three \emph{Direct} methods collapse on AP50 (Pix2Poly $0.102$, PolyWorld $0.016$, RoIPoly $0.001$). HoliTracer (\emph{Rep.}), designed for holistic large-tile inference, transfers poorly to the $512\times512$ patches used in \model{} ($0.034$). A clearer ordering emerges in ring-structure metrics. Hole-F1 is led by ACPV-Net ($0.527$), U-Net + Poly. ($0.514$), and HiSup ($0.473$), while the remaining baselines drop sharply (Mask R-CNN + Poly. $0.167$, GCP $0.015$, FFL $0.000$, and all \emph{Direct} methods below $0.05$). The distribution does not cleanly split along \emph{Exp.}/\emph{Imp.} lines. GCP and FFL are \emph{Exp.} yet recover almost no holes, while the \emph{Imp.} U-Net + Poly. and HiSup rank in the top three. Topo-EM flips the ranking. U-Net + Poly. is best ($0.531$), Mask R-CNN + Poly. follows ($0.519$), and GCP scores only $0.065$ despite strong AP50 and E-BIoU ($0.312$), because it rarely recovers the complete ring structure. Together, these numbers show that region agreement and ring-structure correctness are largely uncorrelated on Inria dataset.

\begin{table*}[t]
\centering
\scriptsize
\caption{
Main 
results across the four single-class polygon generation tasks.
\emph{Building} uses the Inria dataset~\cite{maggiori2017can}; \emph{Road},
\emph{Veg.} (vegetation), and \emph{Unveg.} (unvegetated) use the Deventer dataset~\cite{jiao2026acpv}.
O, E, and H denote the Overall, Ext., and Hole ring-aware scopes. AP50, BIoU,
Hole-F1, and Topo-EM are higher-is-better; POLIS and MTA are lower-is-better.
``--'' marks scores that are undefined because no hole match is accepted on
that task. Best value per task and per column is \textbf{bold}.
}
\label{tab:main_benchmark}
\setlength{\tabcolsep}{2.6pt}
\resizebox{\textwidth}{!}{%
\begin{tabular}{llcccccccccccc}
\toprule
 & & \multicolumn{1}{c}{Region} &
 \multicolumn{9}{c}{Vector-boundary quality} &
 \multicolumn{2}{c}{Ring structure} \\
\cmidrule(lr){3-3}\cmidrule(lr){4-12}\cmidrule(lr){13-14}
Method & Task & AP50 &
\multicolumn{3}{c}{BIoU} &
\multicolumn{3}{c}{POLIS} &
\multicolumn{3}{c}{MTA} &
Hole-F1 & Topo-EM \\
\cmidrule(lr){4-6}\cmidrule(lr){7-9}\cmidrule(lr){10-12}
 & & $\uparrow$ & O$\uparrow$ & E$\uparrow$ & H$\uparrow$
 & O$\downarrow$ & E$\downarrow$ & H$\downarrow$
 & O$\downarrow$ & E$\downarrow$ & H$\downarrow$
 & $\uparrow$ & $\uparrow$ \\
\midrule
U-Net + Poly.       & Building & 0.670 & \textbf{0.294} & 0.308 & \textbf{0.234} & 2.62 & 2.99 & 2.79 & 47.52 & 47.02 & 48.23 & 0.514 & \textbf{0.531} \\
                    & Road     & \textbf{0.520} & \textbf{0.328} & \textbf{0.311} & \textbf{0.337} & \textbf{5.16} & 8.44 & 2.66 & 52.66 & 52.58 & 46.93 & \textbf{0.466} & \textbf{0.144} \\
                    & Veg.     & \textbf{0.382} & 0.340 & 0.313 & 0.208 & \textbf{4.41} & 6.12 & 5.17 & 50.52 & 49.48 & 48.83 & \textbf{0.022} & 0.222 \\
                    & Unveg.   & \textbf{0.247} & 0.222 & 0.206 & \textbf{0.253} & 5.39 & 8.24 & \textbf{3.15} & 56.70 & 55.43 & 50.68 & \textbf{0.428} & 0.088 \\
Mask R-CNN + Poly.  & Building & \textbf{0.720} & 0.267 & 0.292 & 0.155 & \textbf{2.61} & \textbf{2.72} & 3.78 & 43.10 & 42.49 & 53.40 & 0.167 & 0.519 \\
                    & Road     & 0.113 & 0.234 & 0.273 & 0.184 & 7.75 & 8.90 & 4.28 & 52.06 & 49.33 & 48.81 & 0.010 & 0.081 \\
                    & Veg.     & 0.357 & 0.321 & 0.320 & \textbf{0.244} & 4.50 & \textbf{4.72} & 2.59 & 46.63 & 45.94 & 42.75 & 0.017 & \textbf{0.273} \\
                    & Unveg.   & 0.181 & 0.220 & \textbf{0.234} & 0.186 & \textbf{5.23} & \textbf{6.21} & 3.26 & 53.70 & 51.55 & 53.29 & 0.254 & \textbf{0.099} \\
\midrule
SAM2 + Poly.        & Building & 0.431 & 0.242 & 0.269 & 0.154 & 3.05 & 3.17 & 3.61 & 46.43 & 45.63 & 48.85 & 0.017 & 0.377 \\
                    & Road     & 0.023 & 0.299 & 0.301 & 0.177 & 6.31 & 10.40 & 4.55 & 52.55 & 52.30 & 52.36 & 0.014 & 0.012 \\
                    & Veg.     & 0.079 & 0.309 & 0.298 & 0.243 & 6.06 & 6.43 & 2.73 & 50.19 & 49.54 & 50.46 & 0.003 & 0.099 \\
                    & Unveg.   & 0.005 & \textbf{0.239} & 0.230 & 0.165 & 5.57 & 7.44 & 4.62 & \textbf{51.89} & 50.27 & \textbf{44.06} & 0.009 & 0.020 \\
\midrule
HiSup               & Building & 0.496 & 0.282 & 0.289 & 0.224 & 2.84 & 3.79 & 3.02 & 39.82 & 39.66 & 41.43 & 0.473 & 0.387 \\
                    & Road     & 0.389 & 0.267 & 0.257 & 0.256 & 6.49 & 9.40 & 3.97 & 49.84 & 48.74 & \textbf{43.82} & 0.303 & 0.137 \\
                    & Veg.     & 0.264 & 0.295 & 0.270 & 0.176 & 6.86 & 8.88 & 4.07 & 45.40 & 44.46 & 49.65 & 0.016 & 0.181 \\
                    & Unveg.   & 0.110 & 0.181 & 0.180 & 0.207 & 7.66 & 12.31 & 3.85 & 54.24 & 51.16 & 46.72 & 0.350 & 0.044 \\
ACPV-Net            & Building & 0.575 & 0.259 & 0.272 & 0.234 & 2.76 & 3.33 & \textbf{2.64} & 45.76 & 45.37 & 45.84 & \textbf{0.527} & 0.440 \\
                    & Road     & 0.361 & 0.302 & 0.281 & 0.283 & 6.35 & 9.70 & 2.52 & 49.00 & 48.89 & 45.98 & 0.394 & 0.137 \\
                    & Veg.     & 0.274 & 0.332 & 0.306 & 0.153 & 6.24 & 8.27 & 16.81 & 47.03 & 45.70 & 38.23 & 0.006 & 0.154 \\
                    & Unveg.   & 0.082 & 0.157 & 0.169 & 0.173 & 6.94 & 10.90 & 4.07 & 55.84 & 53.31 & 50.77 & 0.265 & 0.026 \\
FFL                 & Building & 0.400 & 0.238 & 0.255 & --    & 3.72 & 4.70 & --   & 45.44 & 44.74 & --    & 0.000 & 0.222 \\
                    & Road     & 0.163 & 0.225 & 0.216 & --    & 10.06 & 15.14 & -- & 56.23 & 56.04 & --    & 0.000 & 0.027 \\
                    & Veg.     & 0.254 & 0.303 & 0.271 & --    & 7.00 & 9.18 & --   & 48.89 & 48.24 & --    & 0.000 & 0.132 \\
                    & Unveg.   & 0.011 & 0.092 & 0.166 & --    & 21.09 & 10.03 & -- & 60.33 & 53.41 & --    & 0.000 & 0.002 \\
GCP                 & Building & 0.631 & 0.288 & \textbf{0.312} & 0.152 & 2.80 & 3.28 & 5.61 & 34.94 & 34.83 & 44.16 & 0.015 & 0.065 \\
                    & Road     & 0.281 & 0.246 & 0.254 & 0.203 & 7.95 & 12.18 & 4.94 & 49.87 & 47.63 & 45.64 & 0.015 & 0.004 \\
                    & Veg.     & 0.203 & 0.313 & 0.299 & 0.184 & 5.72 & 7.78 & 3.75 & 47.31 & 46.28 & 45.76 & 0.001 & 0.024 \\
                    & Unveg.   & 0.068 & 0.134 & 0.179 & 0.151 & 10.64 & 11.78 & 5.56 & 54.85 & 52.66 & 47.60 & 0.009 & 0.008 \\
HoliTracer          & Building & 0.034 & 0.146 & 0.178 & 0.148 & 5.36 & 6.73 & 4.08 & 46.83 & 45.26 & 44.41 & 0.008 & 0.056 \\
                    & Road     & 0.008 & 0.102 & 0.160 & 0.168 & 12.84 & 10.37 & 4.11 & 56.63 & 48.73 & 49.75 & 0.024 & 0.010 \\
                    & Veg.     & 0.097 & 0.256 & 0.248 & 0.218 & 7.66 & 10.31 & 2.49 & 48.21 & 46.61 & \textbf{37.42} & 0.009 & 0.104 \\
                    & Unveg.   & 0.011 & 0.094 & 0.148 & 0.142 & 11.87 & 10.70 & 4.23 & 57.71 & 53.47 & 47.36 & 0.065 & 0.007 \\
\midrule
Pix2Poly            & Building & 0.102 & 0.125 & 0.183 & 0.158 & 5.90 & 7.76 & 4.75 & \textbf{33.04} & \textbf{32.56} & \textbf{36.71} & 0.047 & 0.084 \\
                    & Road     & 0.001 & 0.232 & 0.205 & --    & 7.84 & 15.61 & --   & \textbf{41.68} & \textbf{41.30} & --    & 0.000 & 0.003 \\
                    & Veg.     & 0.028 & \textbf{0.414} & \textbf{0.345} & -- & 41.76 & 63.63 & -- & \textbf{38.11} & \textbf{37.12} & -- & 0.000 & 0.023 \\
                    & Unveg.   & 0.000 & 0.171 & 0.156 & --    & 28.60 & 39.12 & --  & 59.53 & 52.22 & --    & 0.000 & 0.000 \\
PolyWorld           & Building & 0.016 & 0.151 & 0.180 & 0.129 & 3.96 & 7.79 & 5.16 & 41.39 & 40.91 & --    & 0.002 & 0.021 \\
                    & Road     & 0.000 & --    & 0.106 & --    & --    & 7.81 & --   & --    & --    & --    & 0.000 & 0.000 \\
                    & Veg.     & 0.004 & 0.113 & 0.146 & --    & 53.56 & 72.95 & --  & 54.06 & 53.34 & --    & 0.000 & 0.004 \\
                    & Unveg.   & 0.000 & 0.114 & 0.142 & --    & 57.77 & 94.82 & --  & 57.38 & 52.57 & --    & 0.000 & 0.000 \\
RoIPoly             & Building & 0.001 & 0.062 & 0.135 & 0.121 & 9.83 & 10.84 & 7.18 & 52.59 & 47.99 & 44.80 & 0.002 & 0.001 \\
                    & Road     & 0.002 & 0.186 & 0.221 & 0.261 & 9.27 & \textbf{6.81} & \textbf{2.40} & 47.27 & 43.37 & 49.58 & 0.192 & 0.009 \\
                    & Veg.     & 0.032 & 0.196 & 0.217 & 0.210 & 8.67 & 6.46 & \textbf{2.22} & 50.12 & 46.02 & 58.45 & 0.009 & 0.060 \\
                    & Unveg.   & 0.005 & 0.104 & 0.175 & 0.230 & 10.69 & 6.71 & 3.57 & 55.25 & \textbf{49.37} & 45.46 & 0.127 & 0.008 \\
\bottomrule
\end{tabular}}
\end{table*}

\textbf{Results on the Deventer multi-class dataset.}
The Deventer road, vegetation, and unvegetated tasks are substantially harder. The best AP50 drops to $0.520$ (U-Net + Poly., Road), $0.382$ (U-Net + Poly., Vegetation), and $0.247$ (U-Net + Poly., Unvegetated). The relative ordering across method types is similar, with \emph{Seg.} leading on AP50, \emph{Rep.} methods (HiSup, ACPV-Net) close behind on Road and Vegetation, and \emph{Direct} methods again near zero. Deventer Road dataset has, on average, $9.37$ holes per complex instance (Table~\ref{tab:dataset_complexity}), the highest hole density in the benchmark, and therefore provides the most informative signal on hole recovery. U-Net + Poly. reaches Hole-F1 $=0.466$, ACPV-Net $0.394$, and HiSup $0.303$, while the rest of the baselines stay below $0.03$. Topo-EM on Road remains below $0.15$ for every method, indicating that recovering some holes is very different from recovering the complete ring structure. On Vegetation, Pix2Poly attains the highest O-BIoU ($0.414$) and E-BIoU ($0.345$) of any method, but POLIS $=41.76$ and Topo-EM $=0.023$ reveal that its matched polygons are crisp in the boundary buffer yet poorly localized at the vertex level. This error pattern is hidden by exterior-only evaluation. Hole metrics marked ``--'' for FFL, Pix2Poly, and PolyWorld indicate that, although their architectures can represent interior rings and we trained each on the same splits, the trained models rarely produce an accepted hole match.

Overall, three findings can be concluded. First, \emph{Seg.} pipelines lead the \model{} because their region-first training signal transfers robustly across complexity regimes, not because their vector outputs are higher quality. Second, \emph{Direct} decoders that excel on simple-building benchmarks collapse once holes and long, irregular contours dominate the dataset. Third, Topo-EM does not saturate even when AP50 and exterior-boundary scores are strong, so ring-structure correctness is a distinct aspect that current methods do not address.

\subsection{Hole-Aware Evaluation}

\begin{figure*}[t]
\begin{minipage}[t]{0.49\textwidth}
  \centering
  \includegraphics[width=\textwidth]{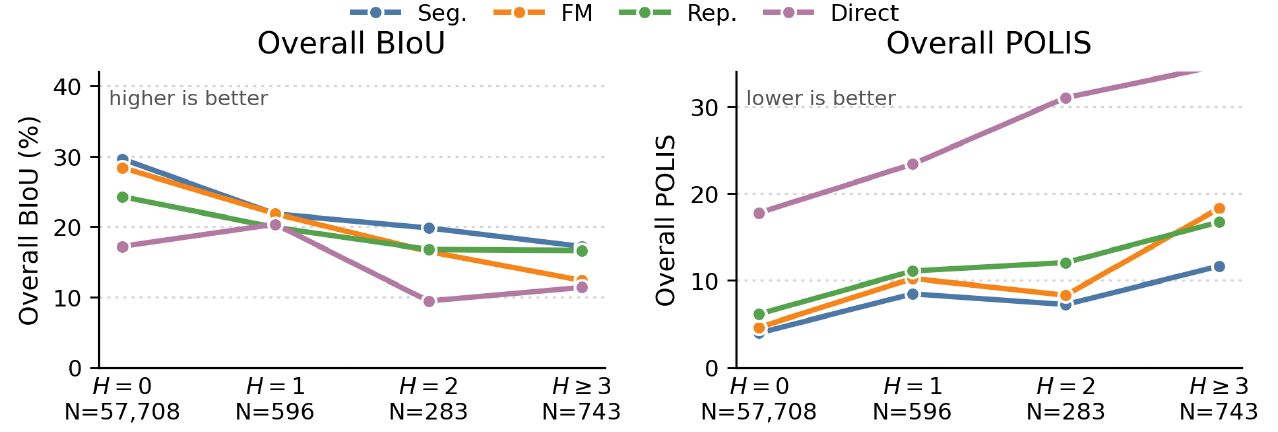}
  \caption{
  Simple-to-complex generalization results. Ground-truth instances are grouped by hole count $H$ with sample size $N$. Each curve reports the average of one method type (\emph{Seg.}, \emph{FM}, \emph{Rep.}, \emph{Direct}) over its constituent baselines.  }
  \label{fig:simple_complex_overall_geometry}
\end{minipage}\hfill
\begin{minipage}[t]{0.49\textwidth}
  \centering
  \includegraphics[width=\textwidth]{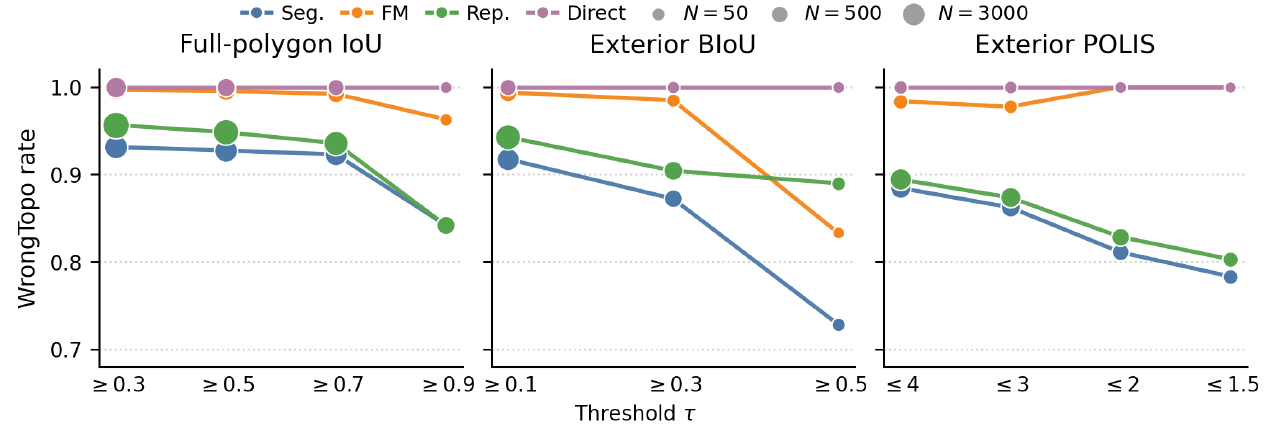}
  \caption{
WrongTopo rate under different geometry filters on complex polygons. From left to right, we filter matched pairs by full-polygon IoU, exterior BIoU, and exterior POLIS; marker size is the number of ground-truth-prediction pairs ($N$). }
  \label{fig:wrong_topo_geometry}

\end{minipage}

\end{figure*}

\textbf{Simple-to-complex generalization.}
We group ground-truth polygon annotations by hole count ($H=0$ to $H\geq3$, where $H$ is a short name for $H(G)$) and report the average Overall BIoU and Overall POLIS of each method type in Figure~\ref{fig:simple_complex_overall_geometry}. Overall BIoU falls, and Overall POLIS grows monotonically with $H$ for every
method type. \emph{Seg.} drops from Overall BIoU $0.296$ at $H=0$ to $0.172$ at $H\geq3$, with Overall POLIS rising from $4.0$ to $11.7$. \emph{FM} degrades more sharply, from $0.284$ to $0.124$ on BIoU and from $4.6$ to $18.3$ on POLIS. The same trend is observed in specialized vector polygon generation methods (both \emph{Rep.} and \emph{Direct.}). In a nutshell, when the complexity increases, the model performance drops significantly.


\textbf{Good exterior geometry still hides topology errors.}
Next, we ask whether a prediction has the correct topology given that its exterior geometry is already good. To make this question quantitative, we define WrongTopo. Let $\mathcal{M}_{\mathrm{complex}}$ be the set of IoU-matched pairs $(\hat{G},G)$ whose ground-truth polygon has at least one hole, and let
$T(\hat{G},G)\in\{0,1\}$ be the per-pair topology indicator defined for Topo-EM in Section~\ref{sec:method_metrics}. Let $q$ denote a geometry filter on matched pairs, chosen from $\{\mathrm{IoU}\geq\tau,\; E\text{-BIoU}\geq\tau,\; E\text{-POLIS}\leq\tau\}$, where the prefix $E$ restricts the score to the exterior ring. $\mathcal{S}_{q,\tau}\subseteq\mathcal{M}_{\mathrm{complex}}$ is the subset of pairs that satisfy $q$ at threshold $\tau$, WrongTopo averages $1-T$ over this subset:
\begin{equation}
    \operatorname{WrongTopo}(q,\tau)
    = \frac{1}{|\mathcal{S}_{q,\tau}|}
      \sum_{(\hat{G},G)\in\mathcal{S}_{q,\tau}}
      \left(1 - T(\hat{G},G)\right).
\end{equation}
$\mathrm{IoU}$ filters on full-polygon overlap, E-BIoU, on exterior-boundary
overlap, and E-POLIS on exterior-boundary distance. A lower WrongTopo score indicates a better topolopy preservation ability.

Figure~\ref{fig:wrong_topo_geometry} shows that no geometry filter removes topology errors. Even at $\mathrm{IoU}\geq0.9$, WrongTopo stays between $0.84$ (\emph{Seg.} and \emph{Rep.}) and $1.00$ (\emph{Direct}), with \emph{FM} at $0.96$. At E-BIoU$\geq0.3$, WrongTopo is $0.87$ for \emph{Seg.}, $0.98$ for \emph{FM}, $0.90$ for \emph{Rep.}, and $1.00$ for \emph{Direct}. At E-POLIS$\leq2$, the rates are $0.81$, $1.00$, $0.83$, and $1.00$, respectively. Good exterior geometry therefore does not tell us whether the interior rings are correct.

\subsection{Qualitative Results}

\begin{figure*}[t]
\centering
\includegraphics[width=\textwidth]{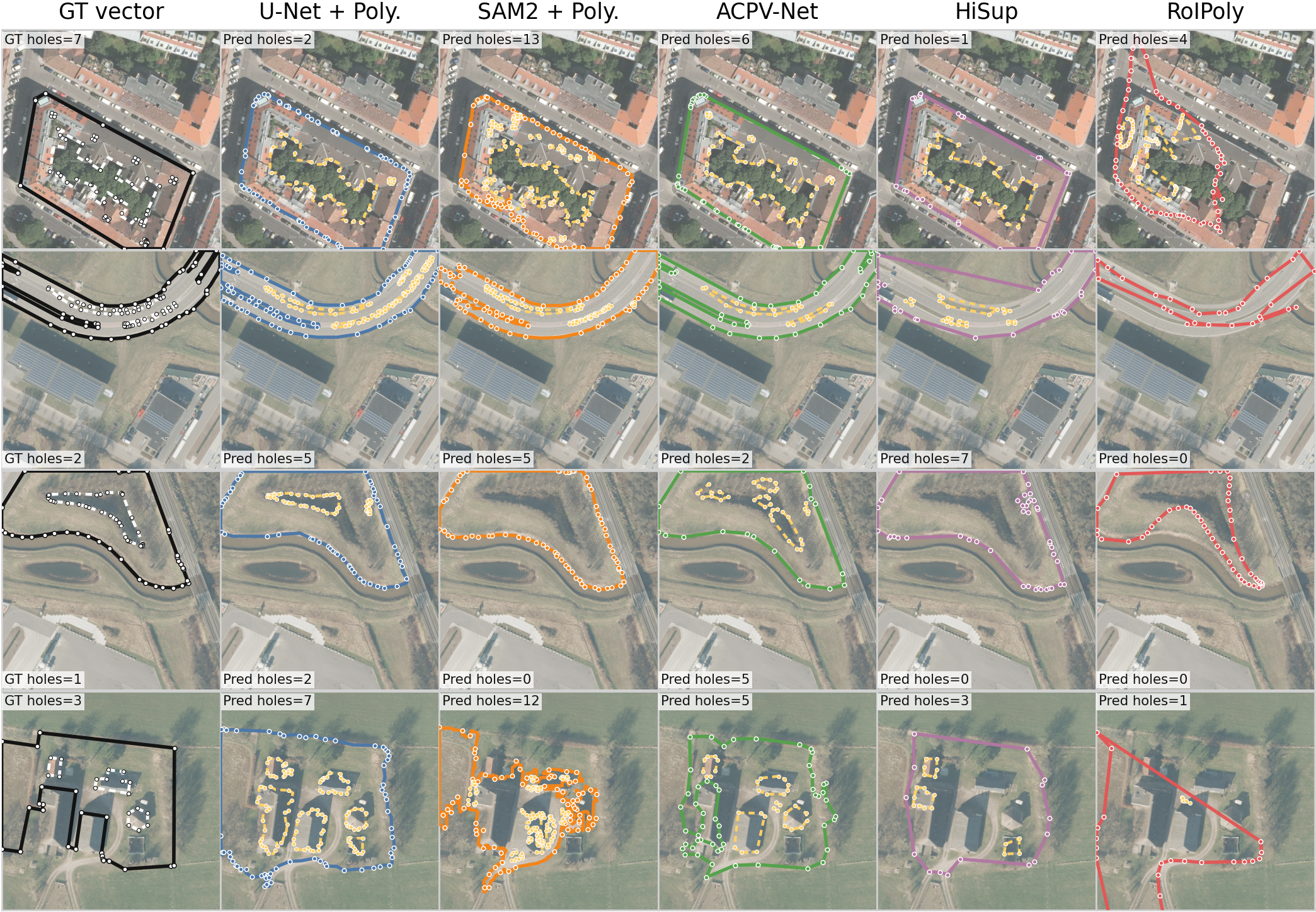}
\caption{
Qualitative vector prediction results. Rows from top to bottom show building, road, vegetation, and unvegetated examples. Columns show five methods, one per type, which are U-Net + Poly. (\emph{Seg.}), SAM2 + Poly. (\emph{FM}), ACPV-Net~\cite{jiao2026acpv} (\emph{Rep.}, \emph{Exp.}), HiSup~\cite{xu2023hisup} (\emph{Rep.}, \emph{Imp.}), and RoIPoly \cite{jiao2025roipoly} (\emph{Direct}, \emph{Imp.})). Each patch is annotated with its ground-truth or predicted hole count. Ground-truth exterior rings are black, interior is white, predicted exterior rings use per-method colors, holes are dashed yellow. Vertices are marked with small circles. See  Appendix~\ref{app:additional_qualitative} for more.
}

\label{fig:qual_simple_complex_vectors}
\end{figure*}



Figure~\ref{fig:qual_simple_complex_vectors} shows that visual quality is often limited by ring topology rather than exterior geometry. Across all four categories, the predicted exterior rings usually follow the target object, but the interior rings are wrongly represented. The building and road examples contain many ground-truth holes, yet different methods recover very different hole counts; the vegetation and unvegetated examples further show that even plausible outer boundaries can have wrong or incomplete holes. These patterns match the quantitative results: \emph{Seg.} and \emph{FM} pipelines often turn mask fragments into false holes, while \emph{Rep.} and \emph{Direct} methods produce smoother contours but still fail to recover reliable interior-ring structure.

\subsection{Why Do Baselines Fail?}
\label{sec:why_fail}

We further conduct baseline-setting ablations in Appendix~\ref{app:method_setting_ablation}. The results show that the eleven methods fail for different reasons, and none of the tested settings produces reliable interior rings. \emph{Seg.} and \emph{FM} pipelines rely on binary foreground masks without explicit interior-ring supervision, so holes lost in the mask cannot be recovered by polygonization; for example, SAM2 with ground-truth boxes reaches AP50 $=0.824$ but Hole-F1 remains $0.000$. \emph{Rep.} methods improve exterior geometry but rarely preserve topology: GCP obtains the highest E-BIoU $=0.312$ on Inria, while Topo-EM stays at $0.065$. \emph{Direct} methods are limited by upstream proposal or vertex modules, which prevent the decoder from forming valid ring structures; RoIPoly's Hole-F1 drops from $0.310$ to $0.002$ when ground-truth boxes are replaced by Sparse R-CNN proposals~\cite{sun2021sparse}, and Pix2Poly improves from AP50 $=0.000$ to $0.102$ when the maximum vertex length increases from $192$ to $224$, but Hole-F1 only reaches $0.047$. HoliTracer and PolyWorld also remain weak on interior-ring recovery, with accepted hole matches appearing only rarely and not translating into reliable topology exactness. Overall, no configuration achieves Topo-EM above $0.377$, and most remain below $0.1$, indicating that existing polygon generators are still optimized mainly for regions or exterior boundaries. Future models should directly supervise interior rings and their topological roles, enforce polygon validity and ring containment during decoding, and jointly train polygon decoders with their proposal modules rather than relying on separately trained upstream components.

\section{Limitation and Broader Impact}  \label{sec:limitation}
\model{} is limited to four single-class tasks from two aerial-image datasets, and its annotations may inherit OSM incompleteness, raster-label errors, and upstream Deventer label noise. We only include baselines with public code; therefore LLM-based methods such as VectorLLM~\cite{zhang2026vectorllm} are not covered because their code was unavailable at submission time. The benchmark supports multiple geospatial applications such as urban planning and disaster response, but could also be misused for privacy-sensitive mapping and surveillance.

\section{Conclusion}  \label{conclusion}
We presented \model{}, a benchmark and evaluation framework for complex vector polygon generation from remote-sensing imagery. \model{} provides four single-class tasks, dense complex-polygon annotations, 11 public baselines, and ring-aware metrics for evaluating exterior shape and interior topology. Results show that strong region or exterior-boundary scores do not guarantee correct interior rings, and no existing model reliably preserves topology.



\bibliographystyle{plain}
\bibliography{reference}

\appendix
\newpage
\section{Appendix}
\subsection{Evaluation Metric Details}
\label{app:metric_details}

This appendix provides the full definitions of the metrics used in \model{}, organized into the three groups: region agreement, vector-boundary quality, and ring-structure correctness. Each normalized polygon instance has one exterior ring and zero or more interior rings. For a polygon $G$, we write $e(G)$ for its exterior ring and $\mathcal{I}(G)$ for its set of interior rings.

\begin{definition}[Instance Matching]
The predicted polygons $\widehat{\mathcal{P}}(\bI)=\{(\hat{c}_i,\hat{s}_i,\hat{G}_i)\}_{i=1}^{N}$ are first matched to ground-truth instances $\mathcal{P}(\bI)=\{(c_j,G_j)\}_{j=1}^{M}$ using the full polygon IoU. Let $\Omega(G)$ denote the occupied region of polygon $G$, with interior rings subtracted from the exterior-ring region. The instance IoU is
\begin{equation}
    \operatorname{IoU}(\hat{G},G)
    =
    \frac{|\Omega(\hat{G})\cap\Omega(G)|}
         {|\Omega(\hat{G})\cup\Omega(G)|}.
\end{equation}
If $\operatorname{IoU}(\hat{G},G) \geq \beta$, then we can say the predicted polygon $\hat{G}$ \emph{matches} the ground-truth polygon $G$, where $\beta$ denotes the IoU matching threshold. 
Note that, the predicted polygons $\widehat{\mathcal{P}}(\bI)$ are sorted by the model's confidence scores $\{\hat{s}_i\}_{i=1}^{N}$, and each ground-truth instance $(c_j,G_j)$ is matched at most once. We denote the resulting set of matched prediction--ground-truth instance pairs by $\mathcal{M}_{\mathrm{inst}}$, and the unmatched ground-truth and prediction sets by $\mathcal{U}_{\mathrm{gt}}$ and $\mathcal{U}_{\mathrm{pred}}$.
\label{def:instance_match}
\end{definition}

\subsubsection{Region agreement metrics}

\begin{definition}[AP50]
AP50 is the average precision computed from full polygon IoU with matching threshold $\beta = 0.50$ (see Definition \ref{def:instance_match}).
\label{def:ap50}
\end{definition}
\vspace{-0.3cm}
AP50 is reported once per benchmark task. Because AP50 depends only on occupied area, it does not distinguish between errors on the exterior ring, errors on interior rings, and topology failures.

\subsubsection{Vector-boundary quality metrics}

\begin{definition}[Metric Scope]
BIoU, POLIS, and MTA are reported under three ring-aware scopes that restrict which rings of a matched instance pair are used:
\begin{equation}
    \mathcal{R}^{\mathrm{Overall}}(G)=\{e(G)\}\cup\mathcal{I}(G),
    \qquad
    \mathcal{R}^{\mathrm{Ext}}(G)=\{e(G)\},
    \qquad
    \mathcal{R}^{\mathrm{Hole}}(G)=\mathcal{I}(G).
\end{equation}
The \emph{Overall} metric uses all rings of a matched pair, the \emph{Ext.} metric uses the paired exterior rings only, and the \emph{Hole} metric uses interior-ring pairs produced by the role-specific ring matching defined in the next group. Hole-scope metrics therefore measure the \emph{shape quality} of recovered holes. \label{def:metric_scope}
\end{definition}
Missed and hallucinated holes are counted separately by Hole-F1 and Topo-EM, which will be described later.

\begin{definition}[Boundary IoU (BIoU)]
Given a set of rings $\mathcal{R}$, let $B_r(\mathcal{R})$ be the union of $r$-pixel boundary buffers around all rings in $\mathcal{R}$. Boundary IoU is
\begin{equation}
    \operatorname{BIoU}_r(\mathcal{R}_1,\mathcal{R}_2)
    =
    \frac{|B_r(\mathcal{R}_1)\cap B_r(\mathcal{R}_2)|}
         {|B_r(\mathcal{R}_1)\cup B_r(\mathcal{R}_2)|}.
\end{equation}
A higher BIoU indicates a better model performance. 
\label{def:BIoU}
\end{definition}

\begin{definition}[Overall-BIoU, Ext-BIoU, and Hole-BIoU]
Overall-BIoU and Ext-BIoU apply $\operatorname{BIoU}_r$ to $\mathcal{R}^{\mathrm{Overall}}$ and $\mathcal{R}^{\mathrm{Ext}}$ of each matched instance pair. Hole-BIoU is the mean over accepted hole matches:
\begin{equation}
    \operatorname{Hole\mbox{-}BIoU}
    =
    \frac{1}{|\mathcal{M}_{\mathrm{hole}}|}
    \sum_{(\hat{h},h)\in\mathcal{M}_{\mathrm{hole}}}
    \operatorname{BIoU}_r(\{\hat{h}\},\{h\}),
\end{equation}
where $\mathcal{M}_{\mathrm{hole}}$ is defined in Definition \ref{def:match_hole_set} below. If a method produces no accepted hole matches on a task, Hole-BIoU is undefined and reported as ``--''. In that case, Hole-F1 (Definition \ref{def:hole_f1}) reflects the failure to recover holes.
\label{def:oei_BIoU}
\end{definition}

\begin{definition}[POLIS]
POLIS measures the average bidirectional distance from polygon vertices to the opposite boundary. For two ring sets $\mathcal{R}_1,\mathcal{R}_2$ with vertex sets $V_1,V_2$ and boundaries $\partial\mathcal{R}_1,\partial\mathcal{R}_2$,
\begin{equation}
    \operatorname{POLIS}(\mathcal{R}_1,\mathcal{R}_2)
    =
    \frac{1}{2}
    \left(
    \frac{1}{|V_1|}\sum_{v\in V_1} d(v,\partial\mathcal{R}_2)
    +
    \frac{1}{|V_2|}\sum_{v\in V_2} d(v,\partial\mathcal{R}_1)
    \right).
\end{equation}
Here, $\mathcal{R}_1,\mathcal{R}_2$ can denote either exterior polygon rings or interior hole rings. A lower POLIS indicates a better model performance. 
\label{def:POLIS}
\end{definition}
\begin{definition}[Overall-POLIS, Ext-POLIS, and Hole-POLIS]
Overall-POLIS and Ext-POLIS apply POLIS to $\mathcal{R}^{\mathrm{Overall}}$ and $\mathcal{R}^{\mathrm{Ext}}$. Hole-POLIS averages per-pair POLIS over $\mathcal{M}_{\mathrm{hole}}$:
\begin{equation}
    \operatorname{Hole\mbox{-}POLIS}
    =
    \frac{1}{|\mathcal{M}_{\mathrm{hole}}|}
    \sum_{(\hat{h},h)\in\mathcal{M}_{\mathrm{hole}}}
    \operatorname{POLIS}(\{\hat{h}\},\{h\}).
\end{equation}
Hole-POLIS should be read together with Hole-F1 (Definition \ref{def:hole_f1}) because it characterizes the geometric quality of recovered holes, not the completeness of hole recovery.
\label{def:oei_POLIS}
\end{definition}

\begin{definition}[MTA]
MTA measures the tangent-angle discrepancy between predicted and ground-truth polygon contours. We uniformly sample polygon exterior and interior contour segments from the predicted ring set and project each sample to the nearest ground-truth boundary segment. For a sampled predicted segment $\Delta p_\ell$ and its projected ground-truth segment $\Delta q_\ell$, the orientation-invariant tangent-angle error is
\begin{equation}
    \alpha_\ell =
    \arccos
    \left(
    \frac{|\langle \Delta p_\ell,\Delta q_\ell\rangle|}
         {\|\Delta p_\ell\|_2\|\Delta q_\ell\|_2}
    \right).
\end{equation}
Here, $\langle \cdot, \cdot \rangle$ denotes a dot product between two vectors and $\|\cdot\|_2$ denotes the vector L2 norm. 
Let $\mathcal{V}$ be the set of valid projected segments after the precision and stretch filters. Then MTA is computed as
\begin{equation}
    \operatorname{MTA}
    =
    \max_{\ell\in\mathcal{V}}\alpha_\ell .
\end{equation}
A lower MTA indicates a better model performance. 
\label{def:MTA}
\end{definition}

\begin{definition}[Overall-MTA, Ext-MTA, and Hole-MTA]
Overall-MTA and Ext-MTA apply MTA to $\mathcal{R}^{\mathrm{Overall}}$ and $\mathcal{R}^{\mathrm{Ext}}$. Hole-MTA averages over $\mathcal{M}_{\mathrm{hole}}$:
\begin{equation}
    \operatorname{Hole\mbox{-}MTA}
    =
    \frac{1}{|\mathcal{M}_{\mathrm{hole}}|}
    \sum_{(\hat{h},h)\in\mathcal{M}_{\mathrm{hole}}}
    \operatorname{MTA}(\{\hat{h}\},\{h\}).
\end{equation}
Like Hole-BIoU and Hole-POLIS, Hole-MTA is a conditional shape-quality metric for recovered holes and should be interpreted together with Hole-F1.
\label{def:oei_MTA}
\end{definition}

\subsubsection{Ring-structure correctness}

Ring-scoped and topology metrics require a correspondence between predicted and ground-truth rings. Thus, we give the definition of role-specific ring matching below.
\begin{definition}[Role-specific Ring Matching]
For each IoU-matched instance pair $(\hat{G},G)\in\mathcal{M}_{\mathrm{inst}}$, the exterior rings $e(\hat{G})$ and $e(G)$ are paired directly because each normalized polygon has exactly one exterior ring, and the exterior pair is considered matched when $\operatorname{BIoU}_r(\{e(\hat{G})\},\{e(G)\})$ exceeds the exterior matching threshold $\gamma$. Interior rings $\mathcal{I}(\hat{G})$ and $\mathcal{I}(G)$ are matched one-to-one via Hungarian assignment using per-ring BIoU, and a predicted hole and a ground-truth hole form an \emph{accepted hole match} only when their BIoU exceeds the hole matching threshold $\gamma$. 
\label{def:role_ring_math}
\end{definition}

\begin{definition}[Matched Hole Set $\mathcal{M}_{\mathrm{hole}}$]
    Based on Definition \ref{def:role_ring_math}, we denote the set of all accepted hole matches across the whole evaluation set by $\mathcal{M}_{\mathrm{hole}}$. 
\label{def:match_hole_set}
\end{definition}
Unless otherwise stated, the main benchmark uses $r=1$ pixel and $\gamma=0.1$.

\begin{definition}[Hole-F1]
Hole-F1 treats interior rings as detection objects. Let $\mathrm{TP}_{\mathrm{h}} = |\mathcal{M}_{\mathrm{hole}}|$, let $\mathrm{FP}_{\mathrm{h}}$ be the number of unmatched predicted holes (including holes from unmatched predicted instances), and let $\mathrm{FN}_{\mathrm{h}}$ be the number of unmatched ground-truth holes (including holes from unmatched ground-truth instances). Then
\begin{equation}
    \operatorname{Hole\mbox{-}F1}
    =
    \frac{2\mathrm{TP}_{\mathrm{h}}}
         {2\mathrm{TP}_{\mathrm{h}}+\mathrm{FP}_{\mathrm{h}}+\mathrm{FN}_{\mathrm{h}}}.
\end{equation}
\label{def:hole_f1_app}
\end{definition}
A method that emits only exterior rings, therefore receives zero hole recall on hole-bearing instances.

\begin{definition}[Valid Polygon Geometry]
For topology evaluation, we say that a predicted polygon $\hat{G}$ is \emph{valid} if it satisfies the following five requirements: 
\begin{enumerate}[itemsep=1pt, parsep=0pt, topsep=0pt]
    \item Its exterior ring $e(\hat{G})$ is simple;
    \item Each interior ring $\hat{h} \in \mathcal{I}(\hat{G})$ is simple; 
    \item All interior rings $\mathcal{I}(\hat{G})$ lie inside the exterior ring $e(\hat{G})$; 
    \item Interior rings do not overlap or cross one another;
    \item The occupied region is non-empty. 
\end{enumerate}
\label{def:geom_valid}
\end{definition}
Invalid polygons are counted as topology failures.

\begin{definition}[Topo-EM]
Topo-EM is computed over all IoU-matched instance pairs, with unmatched ground-truth instances and unmatched predictions counted as failures. A matched pair receives $T(\hat{G},G)=1$ only when all of the following hold:
\begin{enumerate}[itemsep=1pt, parsep=0pt, topsep=0pt]
    \item $\hat{G}$ is a valid polygon according to Definition \ref{def:geom_valid};
    \item $e(\hat{G})$ is matched to $e(G)$;
    \item The predicted and ground-truth hole counts are equal, i.e., $H(\hat{G}) = H(G)$;
    \item Every ground-truth hole is matched one-to-one with a predicted hole.
\end{enumerate}
Otherwise $T(\hat{G},G)=0$. We compute the Topo-EM as follows:
\begin{equation}
    \operatorname{Topo\mbox{-}EM}
    =
    \frac{1}{K}
    \sum_{(\hat{G},G)\in\mathcal{M}_{\mathrm{inst}}}
    T(\hat{G},G),
    \qquad
    K=
    |\mathcal{M}_{\mathrm{inst}}|
    +|\mathcal{U}_{\mathrm{gt}}|
    +|\mathcal{U}_{\mathrm{pred}}|.
\end{equation}
\label{def:topo_em_app}
\end{definition}
Topo-EM is stricter than AP50 and the exterior-scope boundary metrics -- it penalizes missing holes, hallucinated holes, invalid polygons, and incorrect complete ring structures that can be hidden by region-level overlap.

\subsubsection{How the Metrics Separate Error Types}
\label{app:error_types}
The three metric groups respond to different error types, so the severity of an error is reflected by which metrics it affects. \emph{Instance-level errors}, such as merging neighboring buildings into one polygon, break instance matching (Definition~\ref{def:instance_match}) and therefore reduce AP50 and all downstream metrics. \emph{Ring-level errors}, i.e., missing or extra interior rings, preserve the instance match but reduce Hole-F1 and Topo-EM. \emph{Geometric errors} in recovered rings are measured separately by BIoU, POLIS, and MTA under the Overall, Ext., and Hole scopes. Reading the three groups together therefore distinguishes a severe instance error from a topology error on an otherwise correct instance, and both from a small boundary deviation.

\subsection{Sensitivity Analysis on Evaluation Metrics}
\label{app:metric_sensitivity}
In this section, we analyze whether the proposed metrics are sensitive to two evaluator choices, i.e., the minimum hole area $A_{min}$ used to filter small interior rings, and the ring matching threshold $\gamma$ used to determine whether two rings are matched. The goal is to verify that our main conclusion is not driven by a single threshold setting.

\subsubsection{Metric Sensitivity to Minimum Hole Area $A_{min}$} \label{sec:sensitive_min_hole_area}
\begin{figure}[t]
  \centering
  \includegraphics[width=\linewidth]{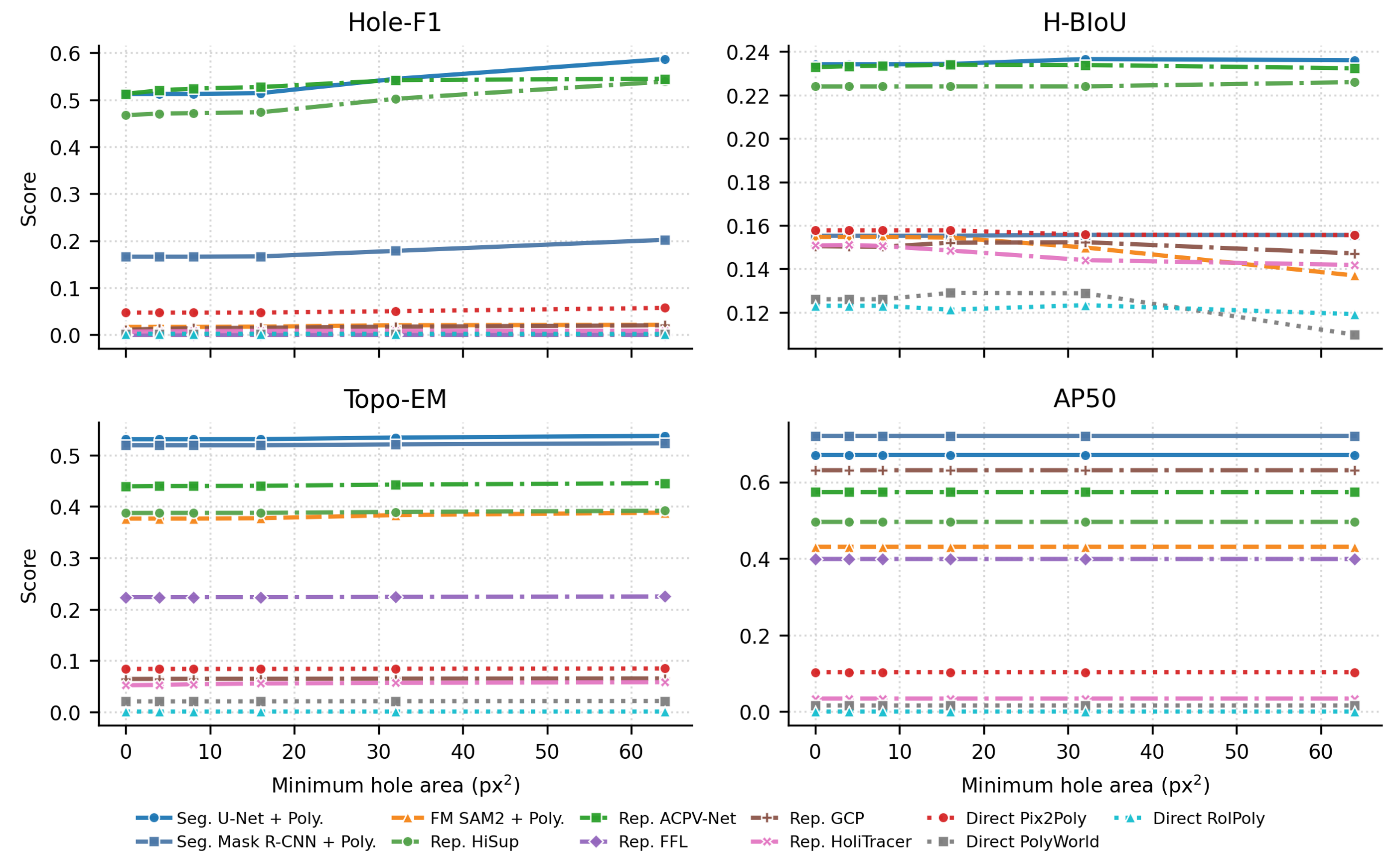}
  \caption{
  Metric sensitivity to the minimum hole area on the Inria Building dataset. We vary the minimum area used to keep ground-truth and predicted interior rings while fixing the ring matching threshold to $\gamma=0.1$. Hole-F1 and H-BIoU are more sensitive to this filtering because they directly evaluate interior rings, while AP50 and Topo-EM remain relatively stable. This shows that the proposed hole-aware metrics are not dominated by a single choice of minimum hole size.
  }
  \label{fig:app_sweep_min_hole_area}
\end{figure}

We first evaluate whether the topology metrics are dominated by many tiny holes. Figure~\ref{fig:app_sweep_min_hole_area} sweeps the minimum hole area $A_{min}$ from $0$ to $64$ pixels (denoted by "px") on the Inria Building dataset, since the Inria dataset has the most complex buildings with diverse polygon complexity. At each threshold, holes smaller than the threshold $A_{min}$ are removed from both ground truth and predictions before evaluation. We evaluate all 11 models on the Inria building dataset under different thresholds $A_{min}$ by using four metrics --  such as $H$-F1, $H$-BoI, Topo-Em, and AP50.

The main trend is stable across different thresholds. Filtering small holes increases Hole-F1 for the baseline, especially U-Net + Poly., ACPV-Net, and HiSup, because the remaining holes are larger and easier to match. However, the relative performance ordering does not change. Methods that fail to recover holes at the default threshold still remain weak after small holes are removed.
Topo-EM is less affected than Hole-F1. This is expected because Topo-EM requires the whole polygon topology to be correct, including exterior matching, the exact number of holes, and one-to-one hole matching. Removing small holes reduces some difficult cases, but it does not fix missing holes, extra holes, or invalid ring structures. AP50 is almost unchanged, which again shows that region overlap is insensitive to interior-ring topology. 

Overall, this analysis shows that our conclusion is not driven only by tiny holes: current methods still struggle with topology even when small holes are filtered out.

\subsubsection{Metric Sensitivity to Ring Matching Threshold $\gamma$} \label{sec:sensitive_ring_match_threshold}
\begin{figure}[t]
  \centering
  \includegraphics[width=\textwidth]{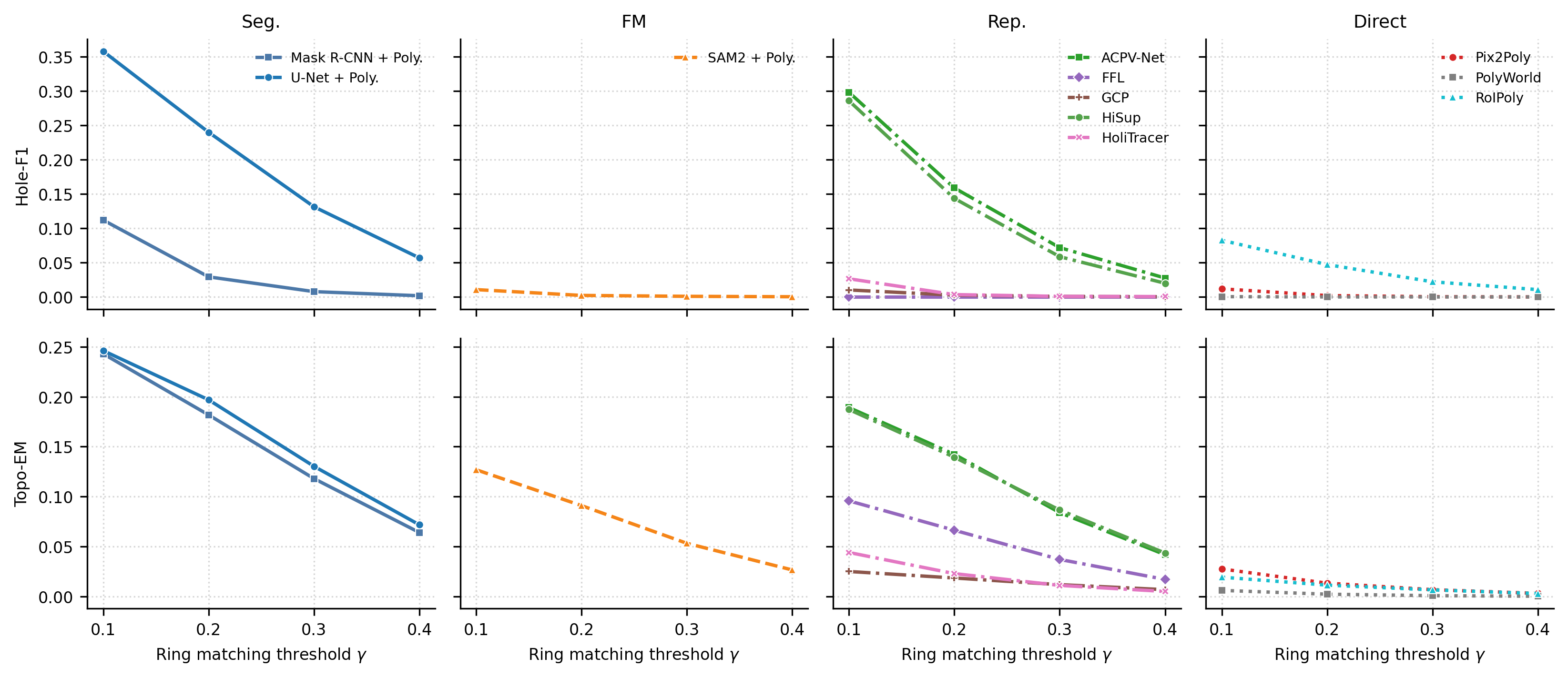}
  \caption{
  Metric sensitivity to the ring matching threshold $\gamma$. Scores are macro-averaged over the four tasks and grouped by method type. Increasing $\gamma$ makes ring matching stricter, which lowers both Hole-F1 and Topo-EM, but the leading methods remain largely stable across thresholds.
  }
  \label{fig:app_ring_threshold_sensitivity}
\end{figure}

Figure~\ref{fig:app_ring_threshold_sensitivity} studies how the proposed ring-structure metrics change when the ring matching threshold $\gamma$ becomes stricter. We report macro-averaged Hole-F1 and Topo-EM over the four tasks and group methods by their taxonomy. As expected, both metrics of all models decrease as $\gamma$ increases because a predicted interior ring must align more closely with the ground-truth ring to be counted as a match.

The main ranking pattern is stable under this sweep. For Hole-F1, U-Net + Poly., ACPV-Net, and HiSup remain the strongest methods across thresholds, while most FM and Direct methods stay close to zero. For Topo-EM, U-Net + Poly. and Mask R-CNN + Poly. remain the strongest Seg. baselines, and HiSup remains the strongest Rep. method. This shows that the low topology scores are not caused by one arbitrary threshold choice of $\gamma$. Stricter matching lowers the absolute scores, but it does not change the conclusion that current baselines still fail to recover the complete interior-ring topology.

\subsubsection{Metric Robustness to Vertex Density} \label{sec:sensitive_vertex_density}
The same polygon shape can be represented with different numbers of vertices, e.g., a straight edge can be stored as two vertices or with additional collinear points. Most \model{} metrics are invariant to this choice by construction because they measure the shape rather than its vertex list. AP50 depends only on the occupied area, and the same shape always has the same area. BIoU buffers each boundary into a band of width $r$ (Definition~\ref{def:role_ring_math}) and compares the two bands, so two vertex lists that trace the same boundary produce the same band and the same score. Hole-F1 and Topo-EM match rings with per-ring BIoU and inherit the same invariance. MTA samples contours at a fixed spacing of $2$ pixels, so vertex density enters only through the positions of the samples along each edge. POLIS is the only metric whose definition depends directly on vertex density, because it averages distances from each vertex to the other boundary, so vertex density acts as a weight. We keep POLIS unchanged because it is a standard metric and preserving its original definition allows direct comparison with prior work.

We verify this property empirically. We take the predictions of four methods on Deventer Road, insert collinear midpoints into every ring to produce $2\times$ and $4\times$ more vertices while preserving exactly the same shapes, and re-run the evaluator. Table~\ref{tab:app_vertex_density} shows that AP50, E-BIoU, Hole-F1, and Topo-EM are identical to four decimal places across all three vertex densities, while POLIS changes by less than $1\%$. MTA changes by at most $4.1\%$ (U-Net + Poly., O-MTA $52.66\rightarrow54.83$ at $4\times$ vertices), which does not alter the method ordering.

\begin{table}[t]
\centering
\small
\setlength{\tabcolsep}{4pt}
\caption{Metric robustness to vertex density on Deventer Road. Collinear midpoints are inserted into every predicted ring to obtain $2\times$ and $4\times$ more vertices without changing the shape. AP50, E-BIoU, Hole-F1, and Topo-EM are identical for all three densities; POLIS (Overall scope) is reported for $1\times$ / $2\times$ / $4\times$ vertices.}
\label{tab:app_vertex_density}
\begin{tabular}{lccccc}
\toprule
Method & AP50 & E-BIoU & Hole-F1 & Topo-EM & POLIS ($1\times$ / $2\times$ / $4\times$) \\
\midrule
U-Net + Poly. & 0.520 & 0.311 & 0.466 & 0.144 & 5.16 / 5.13 / 5.12 \\
HiSup & 0.389 & 0.257 & 0.303 & 0.137 & 6.49 / 6.51 / 6.50 \\
ACPV-Net & 0.361 & 0.281 & 0.394 & 0.137 & 6.35 / 6.31 / 6.29 \\
RoIPoly & 0.002 & 0.221 & 0.192 & 0.009 & 9.27 / 9.32 / 9.32 \\
\bottomrule
\end{tabular}
\end{table}

\subsection{Baseline Implementation Details}
\label{app:baseline_impl}

This section discusses the training and inference settings that produced the
experimental results shown in Table~\ref{tab:main_benchmark}. All baselines are trained per \model{} task under a $512\times512$ crop using their native loss and optimizer, and consume the same train/val split defined by \model{}. Unless otherwise stated, we run each training job on a single GPU and launch different baselines or tasks as independent jobs. The main benchmark runs were executed on a node with four NVIDIA RTX A6000 GPUs, each with $48$GB memory, using eight CPU dataloader workers per job. We fix the random seed for all baselines to ensure a fair comparison. All $11$ baselines are trained with the same multi-ring ground truth, converted into each method's native supervision format, so interior-ring annotations are available to every method during training. Below, we describe the implementation details of each baseline. For every method, we start from its released repository and follow the default implementation unless otherwise stated.

\textbf{Selection of training settings.}
The baselines have substantially different pipelines, and our goal is to let each method reach its best achievable performance on \model{}. We therefore follow a two-step procedure. For every baseline, we first use the official implementation and the training configuration recommended by its authors; this is always our priority. We then examine the resulting performance and run additional ablations that deviate from the official setting only when the results are clearly abnormal, particularly with respect to interior-ring accuracy (Appendix~\ref{app:method_setting_ablation}). The main table reports the best setting identified for each baseline. Initialization follows the same principle. We explicitly ablate initialization for GCP (training from scratch versus transfer learning, Table~\ref{tab:app_gcp_ablation}) and PolyWorld (training from scratch, fine-tuning, and zero-shot evaluation, Table~\ref{tab:app_polyworld_ablation}). For the remaining methods, we do not sweep alternative initialization strategies, because their official implementations already perform as expected and their authors do not recommend alternative initialization settings; for example, the released Pix2Poly implementation is designed to train its ViT encoder from scratch. Given our limited compute budget, we focus additional ablations on the methods and settings where they are most likely to affect performance, especially interior-ring recovery.

\textbf{U-Net + Poly.}~\cite{ronneberger2015u,zorzi2021machine}
We train U-Net with an EfficientNet-B3 encoder (ImageNet pretraining) for $100$ epochs at batch size $32$, learning rate $10^{-4}$, and a combined Dice and cross-entropy loss. Masks are thresholded at $0.5$ and polygonized by Douglas--Peucker simplification at tolerance $1.0$~px.

\textbf{Mask R-CNN + Poly.}~\cite{he2017mask,zorzi2021machine}
We fine-tune Mask R-CNN with a ResNet-50 FPN backbone from COCO-pretrained weights for $100$ epochs at batch size $6$ and learning rate $5\!\times\!10^{-3}$. Predicted masks with detector score $\geq 0.05$ are polygonized with the same procedure as U-Net + Poly.

\textbf{SAM2 + Poly.}~\cite{muhawenayo2026prue}
We use SAM2 without fine-tuning. For each image, we run the Mask R-CNN detector trained above, keep predicted boxes with score $\geq0.5$, and prompt SAM2 with each box. The returned masks are polygonized with the same tolerance and area thresholds as U-Net + Poly. We conduct an ablation study on different prompts used for SAM2. Table~\ref{tab:app_sam2_ablation} shows that Mask R-CNN box prompts are the selected realistic setting for the main benchmark; the ground-truth box prompt is reported only as an upper-bound diagnostic.

\textbf{HiSup.}~\cite{xu2023hisup}
We train HiSup with an HRNet-48 backbone on $512\times512$ image crops at batch size $8$ and base learning rate $10^{-4}$. Attraction-field, vertex, and mask loss weights follow the released configuration.

\textbf{ACPV-Net.}~\cite{jiao2026acpv}
We train ACPV-Net with an HRNet-32 backbone for $4\!\times\!10^{5}$ iterations at batch size $8$ and base learning rate $6\!\times\!10^{-5}$. Other settings follow the released configuration.

\textbf{FFL.}~\cite{girard2021polygonal}
We train Frame Field Learning with its default U-Net backbone for a total budget of $100$ epochs at batch size $13$, polygonization learning rate $10^{-2}$, and learning-rate decay $\gamma=0.99$, and report the checkpoint with the best validation performance, which is reached after $5$ epochs. This checkpoint occurs early because FFL overfits quickly on our data, especially on hole-bearing buildings. The original FFL paper uses early stopping at $15$--$25$ epochs~\cite{girard2021polygonal}; our $100$-epoch budget with best-validation checkpoint selection is therefore more generous than the official recipe.

\textbf{GCP.}~\cite{zhang2025global}
GCP is trained in two stages. Stage~1 trains the segmentation and corner prediction backbone for $24$ epochs at batch size $24$ and learning rate $10^{-4}$. Stage~2 fine-tunes the Transformer contour-regression module for another $24$ epochs under the same batch size and learning rate, initialized from the stage~1 checkpoint. The global collinearity polygonizer keeps the tolerance and vertex length from its released implementation. Table~\ref{tab:app_gcp_ablation} compares training from scratch with transfer from the WHU-Mix checkpoint; training from scratch gives the stronger AP50, Hole-F1, and Topo-EM in this setting, so the main benchmark uses the scratch-trained GCP model.

\textbf{HoliTracer.}~\cite{wang2025holitracer}
HoliTracer is trained in two stages with a Swin-L backbone. The Context Attention segmentation network is trained for $10$ epochs at batch size $7$ and learning rate $10^{-5}$. The Polygon Sequence Tracer is then trained for $10$ epochs at batch size $8$ and learning rate $10^{-3}$, using the stage-1 checkpoint as its backbone. Table~\ref{tab:app_holitracer_ablation} sweeps the corner confidence threshold. The sweep selects corner threshold $0.10$ for the reported HoliTracer setting, but the low Hole-F1 and Topo-EM show that changing this threshold alone does not solve interior-ring recovery.

\textbf{Pix2Poly.}~\cite{adimoolam2025pix2poly}
We train Pix2Poly end-to-end for $200$ epochs at batch size $28$ and learning rate $2\!\times\!10^{-4}$, using the released ViT-based encoder without ImageNet pretraining. Table~\ref{tab:app_pix2poly_ablation} ablates the maximum vertex length $N_{\text{vert}}$ and affine-rotation augmentation. The main benchmark reports $N_{\text{vert}}=224$ without affine rotation, the strongest configuration under these tested settings.

\textbf{PolyWorld.}~\cite{zorzi2022polyworld}
We train PolyWorld from scratch with its R2U-Net backbone for $100$ epochs at batch size $16$ and learning rate $10^{-4}$. The Sinkhorn-style optimal matching and vertex-detection hyperparameters follow the released configuration. The training from scratch setting, the zero-shot from the pretrained checkpoint setting, and the fine-tuning variants are compared in Table~\ref{tab:app_polyworld_ablation}.

\textbf{RoIPoly.}~\cite{jiao2025roipoly}
RoIPoly uses a ResNet-50 backbone with $160$ learnable proposals per image and $64$ corner queries per polygon. We train the polygon head for $1.35\!\times\!10^{5}$ iterations at batch size $8$ and learning rate $2.5\!\times\!10^{-5}$. During model inference, proposals come from a separately trained Sparse R-CNN~\cite{sun2021sparse} detector (ResNet-50 backbone with ImageNet pretraining) rather than ground-truth boxes. The impact of this proposal source on Hole-F1 is characterized in Table~\ref{tab:app_roipoly_ablation}. The main table (Table \ref{tab:main_benchmark}) uses the Sparse R-CNN proposals. The selection of Sparse R-CNN follows the original paper~\cite{jiao2025roipoly}.

\subsection{Additional Qualitative Results}
\label{app:additional_qualitative}

\begin{figure}[t]
\centering
\includegraphics[width=\textwidth]{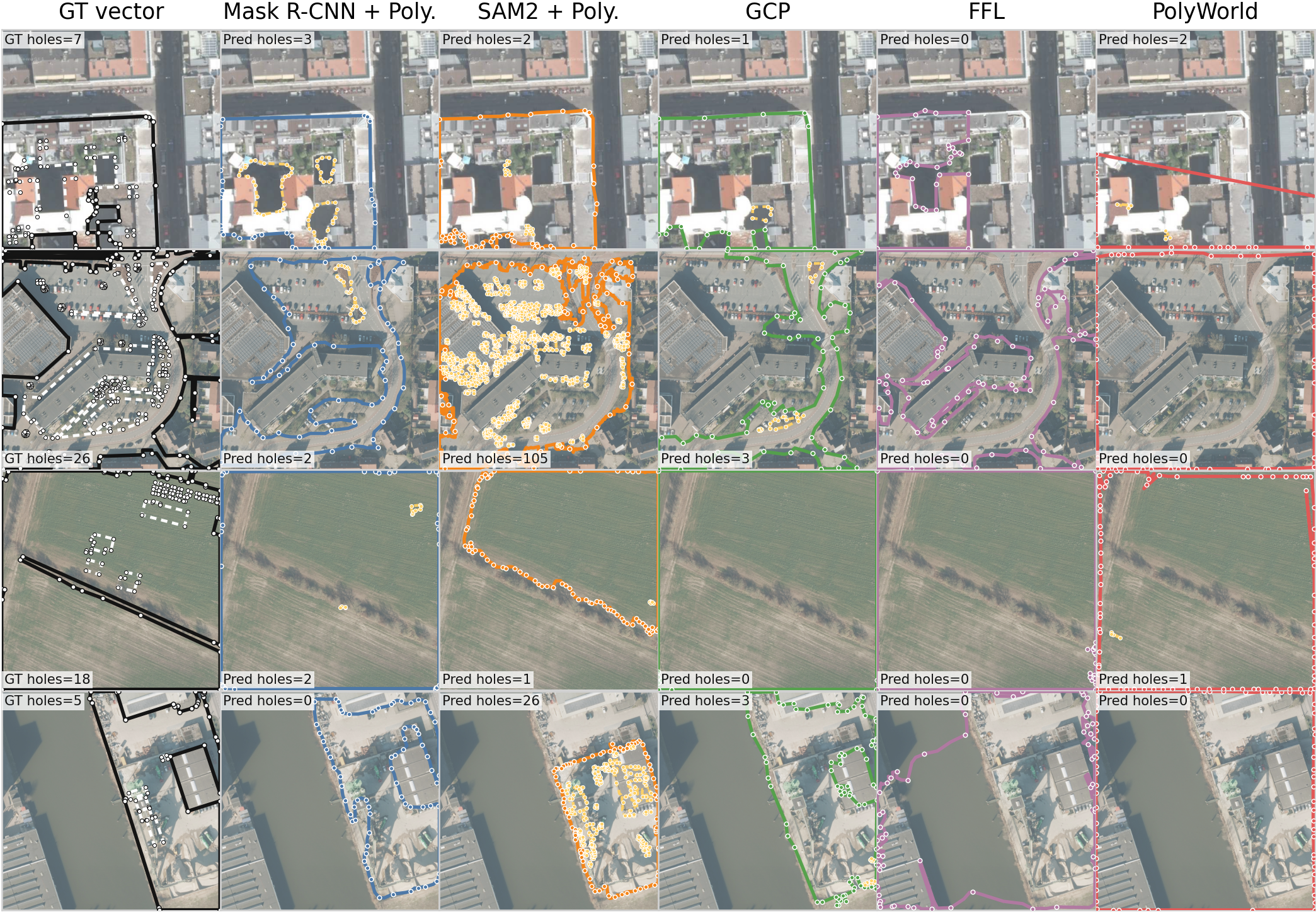}
\caption{Additional qualitative vector predictions. Rows show building, road, vegetation, and unvegetated examples. Columns show the ground-truth vector and five representative methods: Mask R-CNN + Poly. (\emph{Seg.}), SAM2 + Poly. (\emph{FM}), GCP (\emph{Rep.}), FFL (\emph{Rep.}), and PolyWorld (\emph{Direct}). The first column uses black exterior rings for ground truth. Prediction columns show only the predicted vector output: colored solid lines are exterior rings, dashed yellow lines are interior rings, and small circles are vertices. Patch labels report the ground-truth or predicted hole count.}
\label{fig:qual_appendix_01}
\end{figure}

\begin{figure}[t]
\centering
\includegraphics[width=\textwidth]{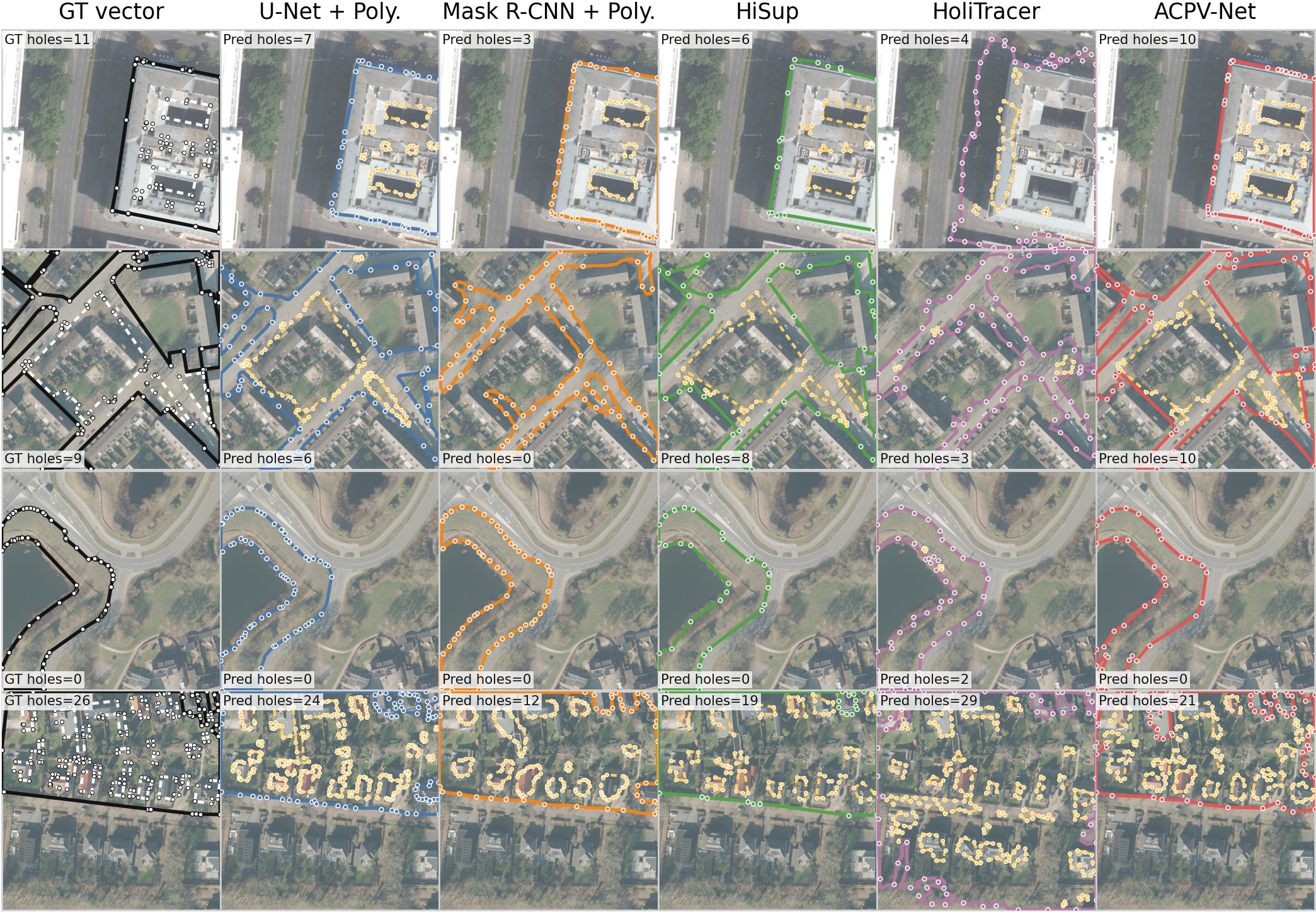}
\caption{Additional qualitative vector predictions with segmentation and representation-to-vector baselines. Rows show the four object categories in the same order as Figure~\ref{fig:qual_appendix_01}. Columns show the ground-truth vector, U-Net + Poly. and Mask R-CNN + Poly. (\emph{Seg.}), HiSup, HoliTracer, and ACPV-Net (\emph{Rep.}). Prediction columns contain no ground-truth overlay, so the figure directly shows each method's predicted exterior ring, interior rings, and vertices.}
\label{fig:qual_appendix_02}
\end{figure}

\begin{figure}[t]
\centering
\includegraphics[width=\textwidth]{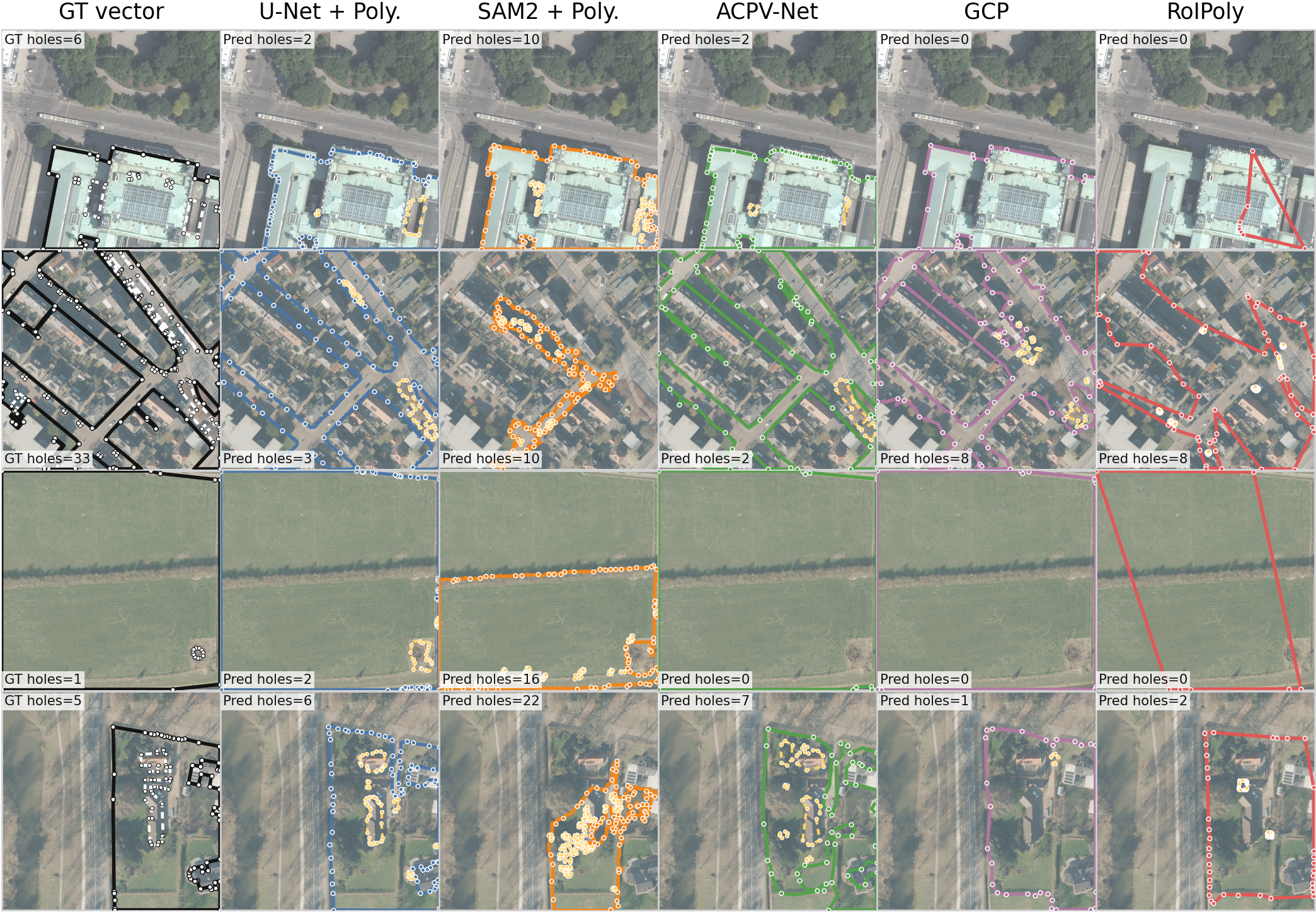}
\caption{Additional qualitative vector predictions comparing all four method types. Columns show the ground-truth vector, U-Net + Poly. (\emph{Seg.}), SAM2 + Poly. (\emph{FM}), ACPV-Net and GCP (\emph{Rep.}), and RoIPoly (\emph{Direct}). Across the four object categories, exterior geometry can remain visually plausible while the number and placement of interior rings differ strongly from the ground truth.}
\label{fig:qual_appendix_03}
\end{figure}

\begin{figure}[t]
\centering
\includegraphics[width=\textwidth]{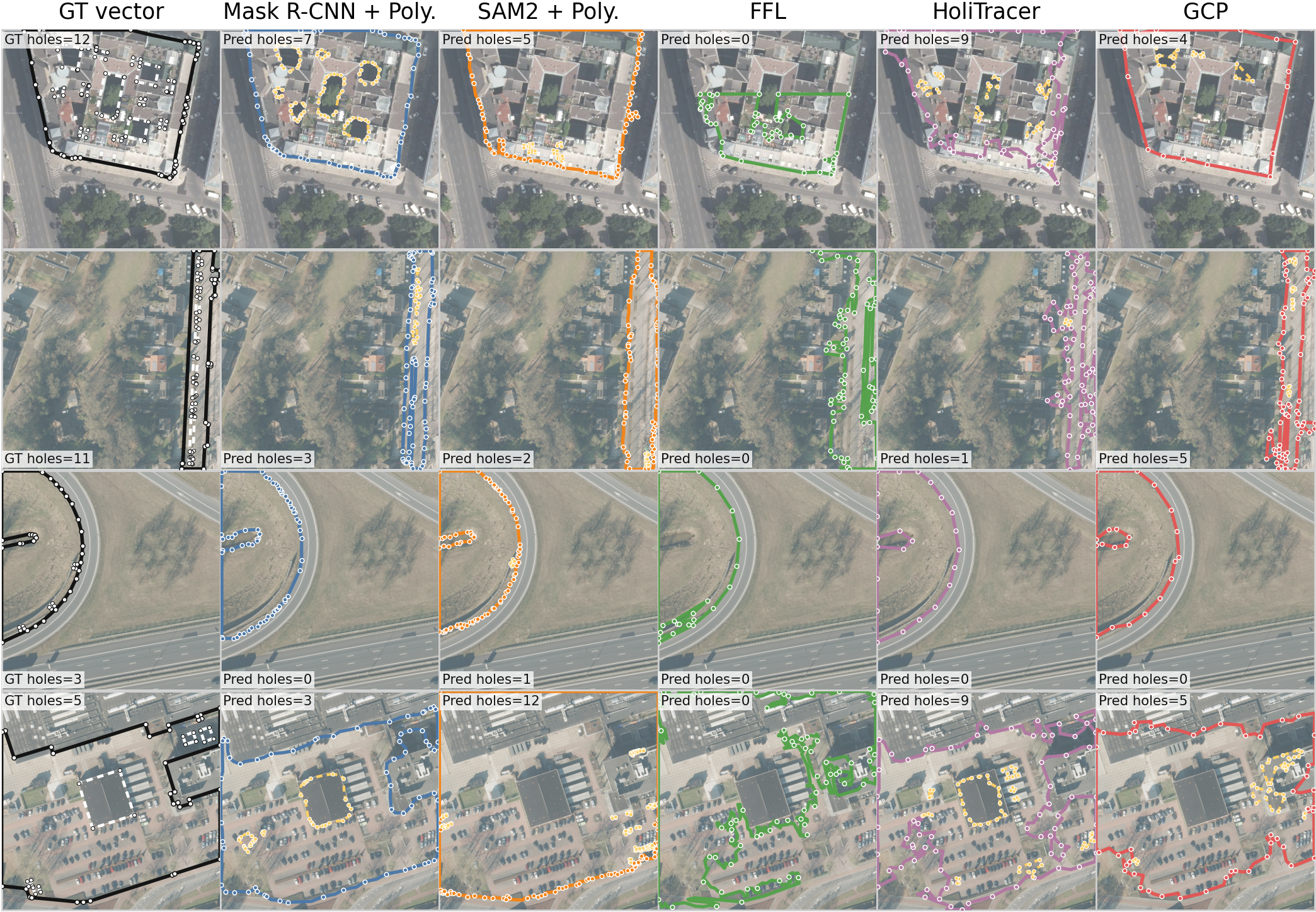}
\caption{Additional qualitative vector predictions emphasizing representation-to-vector methods. Columns show the ground-truth vector, Mask R-CNN + Poly. (\emph{Seg.}), SAM2 + Poly. (\emph{FM}), FFL, HoliTracer, and GCP (\emph{Rep.}). The examples illustrate common topology errors: missing holes, extra holes from fragmented evidence, and exterior rings that trace nearby structures rather than the target instance.}
\label{fig:qual_appendix_04}
\end{figure}

\begin{figure}[t]
\centering
\includegraphics[width=\textwidth]{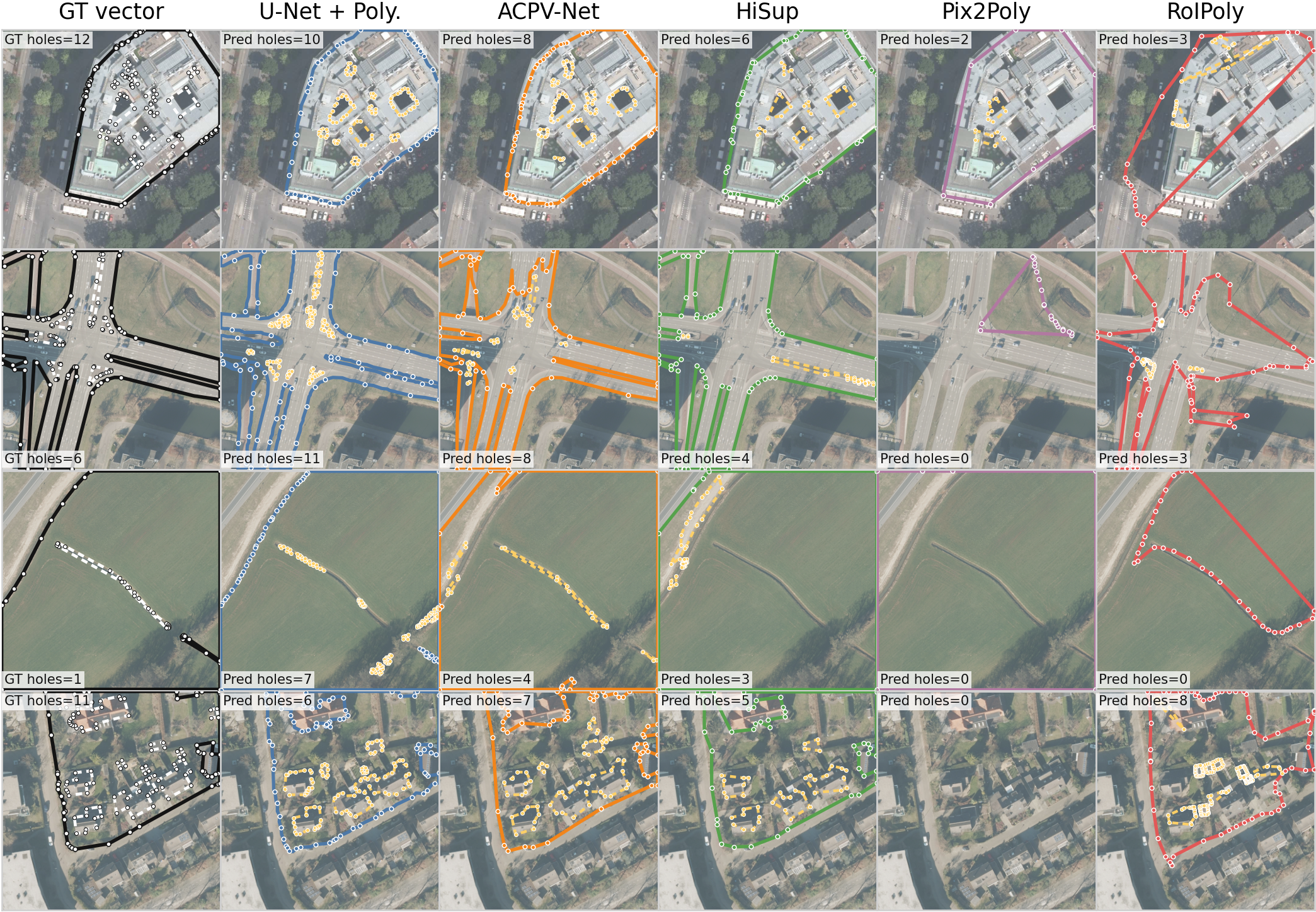}
\caption{Additional qualitative vector predictions including direct vector decoders. Columns show the ground-truth vector, U-Net + Poly. (\emph{Seg.}), ACPV-Net and HiSup (\emph{Rep.}), Pix2Poly and RoIPoly (\emph{Direct}). Direct decoders often produce simple or distorted exterior rings and miss many interior rings, while representation-to-vector methods recover more local structure but still fail to match the complete ring topology.}
\label{fig:qual_appendix_05}
\end{figure}

This section provides additional qualitative examples, as shown in Figure~\ref{fig:qual_appendix_01}--~\ref{fig:qual_appendix_05}. We can observe the same trend as we describe in the main paper.

\subsection{Method Setting Ablation}
\label{app:method_setting_ablation}

This section reports method-level setting ablations used in the discussion. We ablate each baseline under its relevant training or inference settings. All ablation tables in this section use the Inria Building dataset, so the rows compare variants within each method on the same dataset. All values are rounded to three decimals. As shown in Tables~\ref{tab:app_gcp_ablation}--\ref{tab:app_sam2_ablation}, the final setting selected for the main benchmark is highlighted in bold.

\begin{table}[t]
\centering
\small
\setlength{\tabcolsep}{4pt}
\caption{GCP setting ablation on Inria Building. Since GCP releases its checkpoint on WHU-Mix \cite{luo2023diverse}, we ablate the transfer learning setting and the training from scratch setting.}
\label{tab:app_gcp_ablation}
\begin{tabular}{p{0.48\textwidth}cccc}
\toprule
Setting & AP50 & E-BIoU & Hole-F1 & Topo-EM \\
\midrule
Train from the released WHU-Mix checkpoint & 0.190 & 0.588 & 0.000 & 0.004 \\
\textbf{Train from scratch} & \textbf{0.631} & \textbf{0.312} & \textbf{0.015} & \textbf{0.065} \\
\bottomrule
\end{tabular}
\end{table}

\begin{table}[t]
\centering
\small
\setlength{\tabcolsep}{4pt}
\caption{HoliTracer setting ablation on the Inria Building dataset. We sweep the corner confidence threshold used by the vector tracing stage.}
\label{tab:app_holitracer_ablation}
\begin{tabular}{p{0.48\textwidth}cccc}
\toprule
Setting & AP50 & E-BIoU & Hole-F1 & Topo-EM \\
\midrule
Corner threshold $0.00$ & 0.034 & 0.558 & 0.000 & 0.000 \\
\textbf{Corner threshold $0.10$} & \textbf{0.034} & \textbf{0.558} & \textbf{0.000} & \textbf{0.000} \\
Corner threshold $0.50$ & 0.000 & -- & 0.000 & 0.000 \\
\bottomrule
\end{tabular}
\end{table}

\begin{table}[t]
\centering
\small
\setlength{\tabcolsep}{4pt}
\caption{RoIPoly setting ablation on the Inria Building dataset. We compare the polygon head under
ground truth boxes with a setting where a separately trained Sparse R-CNN \cite{sun2021sparse} detector
provides the bounding boxes.}
\label{tab:app_roipoly_ablation}
\begin{tabular}{p{0.18\textwidth}p{0.34\textwidth}cccc}
\toprule
Proposal source & Setting & AP50 & E-BIoU & Hole-F1 & Topo-EM \\
\midrule
Ground truth boxes (upper bound) & Evaluate the RoIPoly polygon head with ground truth object boxes. & 0.158 & 0.680 & 0.310 & 0.082 \\
\textbf{Sparse R-CNN boxes} & \textbf{Train a separate Sparse R-CNN detector on Inria and use its predicted boxes as RoIPoly proposals.} & \textbf{0.001} & \textbf{0.135} & \textbf{0.002} & \textbf{0.001} \\
\bottomrule
\end{tabular}
\end{table}

\begin{table}[t]
\centering
\small
\setlength{\tabcolsep}{4pt}
\caption{Pix2Poly setting ablation on the Inria Building dataset. Pix2Poly predicts vertex tokens with an image-to-sequence Transformer and recovers connectivity with an optimal matching network. We mainly ablate the maximum vertex-sequance length and image data augmentation strategy.}
\label{tab:app_pix2poly_ablation}
\begin{tabular}{p{0.50\textwidth}cccc}
\toprule
Variant & AP50 & E-BIoU & Hole-F1 & Topo-EM \\
\midrule
$512$ input with affine rotation, $192$ maximum vertices & 0.000 & 0.592 & 0.000 & 0.000 \\
$512$ input without affine rotation, $192$ maximum vertices & 0.000 & 0.608 & 0.000 & 0.000 \\
\textbf{$512$ input without affine rotation, $224$ maximum vertices} & \textbf{0.102} & \textbf{0.183} & \textbf{0.047} & \textbf{0.084} \\
\bottomrule
\end{tabular}
\end{table}

\begin{table}[t]
\centering
\small
\setlength{\tabcolsep}{4pt}
\caption{PolyWorld setting ablation on the Inria Building dataset. PolyWorld detects vertex peaks, extracts visual descriptors, and predicts vertex connectivity with a GNN and Sinkhorn-style optimal matching. The variants test whether better vertex supervision or longer fine-tuning fixes topology errors.}
\label{tab:app_polyworld_ablation}
\begin{tabular}{p{0.50\textwidth}cccc}
\toprule
Variant & AP50 & E-BIoU & Hole-F1 & Topo-EM \\
\midrule
Zero-shot using pretrained checkpoint & 0.004 & 0.600 & 0.000 & 0.001 \\
Fine-tuning & 0.012 & 0.544 & 0.000 & 0.000 \\
\textbf{Train from scratch} & \textbf{0.016} & \textbf{0.180} & \textbf{0.002} & \textbf{0.021} \\
\bottomrule
\end{tabular}
\end{table}

\begin{table}[t]
\centering
\small
\setlength{\tabcolsep}{4pt}
\caption{SAM2-assisted polygonization setting ablation on the Inria Building dataset. Each variant changes only the SAM2 prompt source: ground truth boxes, Mask R-CNN predicted boxes, U-Net predicted boxes, U-Net masks with boxes, iterative SAM2 mask refinement, U-Net mask-only prompts, or U-Net box-only prompts. All SAM2 masks are converted to polygons using the same contour extraction and simplification procedure.}
\label{tab:app_sam2_ablation}
\begin{tabular}{p{0.24\textwidth}p{0.34\textwidth}cccc}
\toprule
Prompting variant & Prompt construction & AP50 & O-BIoU & Hole-F1 & Topo-EM \\
\midrule
Ground truth box prompt (upper bound) & Use each ground truth instance box as the SAM2 box prompt. & 0.824 & 0.266 & 0.000 & 0.050 \\ \hline
\textbf{Mask R-CNN box prompt} & \textbf{Use Mask R-CNN predicted boxes as SAM2 box prompts.} & \textbf{0.431} & \textbf{0.242} & \textbf{0.017} & \textbf{0.377} \\
U-Net mask + box & Use a U-Net predicted mask together with its bounding box. & 0.506 & 0.251 & 0.000 & 0.029 \\
Iterative mask refinement & Run SAM2 with U-Net box and mask prompts, then feed the predicted mask back for a second refinement pass. & 0.473 & 0.242 & 0.000 & 0.027 \\
U-Net box only & Use the bounding box of each U-Net predicted instance mask. & 0.402 & 0.243 & 0.000 & 0.024 \\
U-Net mask only & Use the U-Net predicted mask as the only SAM2 prompt. & 0.035 & 0.203 & 0.000 & 0.001 \\
\bottomrule
\end{tabular}
\end{table}

\subsubsection{Failure Analysis by Method Family}
\label{app:failure_analysis}
\begin{figure}[t]
\centering
\includegraphics[width=\textwidth]{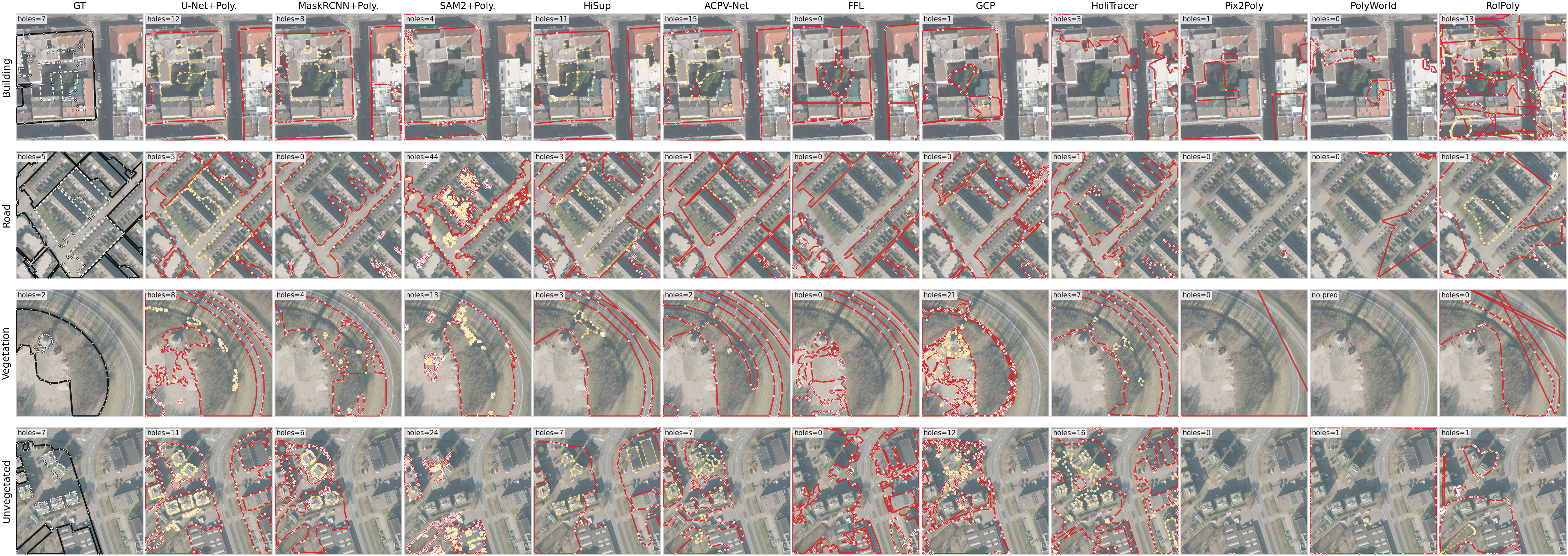}
\caption{Qualitative comparison of all $11$ methods on the same patches. Rows show building, road, vegetation, and unvegetated examples; columns show the ground truth followed by every benchmark method. All methods share one color scheme: black lines are ground-truth exterior rings, red lines are predicted exterior rings, dashed lines are interior rings, and white dots are vertices. Each panel reports its ground-truth or predicted hole count, and ``no pred'' means the method produced no output overlapping the patch. \emph{Seg.} and \emph{Rep.} methods such as U-Net + Poly., HiSup, and ACPV-Net generally recover exterior boundaries and some interior rings, but the predicted hole counts and locations often differ from the ground truth. SAM2 + Poly. tends to hallucinate holes (e.g., $44$ predicted holes on the road example with $5$ ground-truth holes). The direct methods frequently produce fragmented or incomplete instances and almost no interior rings, consistent with their low AP50.}
\label{fig:app_all_methods_qualitative}
\end{figure}

All baselines are trained with the same multi-ring ground truth (Appendix~\ref{app:baseline_impl}), so their low interior-ring scores arise from their polygon representations rather than from missing ring annotations. Figure~\ref{fig:app_all_methods_qualitative} compares all $11$ methods on the same examples, and we analyze the failure modes of each method family below.

\textbf{Segmentation-based methods.} A hole is preserved only if the predicted mask retains the enclosed background. The smoothness bias of segmentation models often fills small holes, which cannot be recovered by the subsequent polygonization.

\textbf{Representation-based methods.} These methods predict attraction fields, frame fields, or contours and then polygonize them. Both stages primarily emphasize object boundaries, so exterior rings are recovered more reliably than interior rings. For example, GCP's collinearity-based simplification yields a Hole-F1 of only $0.015$ on Inria (Table~\ref{tab:app_gcp_ablation}).

\textbf{Direct vector decoders.} On \model{}, the direct methods fail at a more basic level than hole recovery: they do not produce enough complete instances, so their scores collapse already at the region level (AP50), before the ring-structure metrics apply. This differs from their original benchmarks, where each image contains a few simple, compact buildings. AP50 requires complete polygons at IoU $\geq 0.5$ together with a usable confidence ranking, and each direct method breaks one of these requirements.
\begin{itemize}[itemsep=1pt, parsep=0pt, topsep=0pt]
    \item \emph{Pix2Poly} generates all polygons of a patch as one token sequence with a fixed maximum vertex budget. In the original paper, this budget only needs to cover exterior corner points of a few compact buildings in small crops. Each $512\times512$ \model{} patch contains many instances, and the model must also emit vertices on interior rings, which the original task never required. Increasing the vertex budget raises AP50 only from $0.000$ to $0.102$ (Table~\ref{tab:app_pix2poly_ablation}).
    \item \emph{PolyWorld} detects corner points and links them into closed rings through vertex-permutation cycles. Although this can in principle represent holes, the model has no explicit notion of ring role or hierarchy, and it must detect small, low-contrast interior vertices and connect them into separate cycles. On our complex polygons it misses too many corners, so rings do not close properly, the output becomes fragmented, and AP cannot rank good polygons above fragments (Section~\ref{sec:method_protocol}).
    \item \emph{RoIPoly} is a two-stage method that first proposes bounding boxes with an object detector and then decodes one ring per proposal. With ground-truth boxes it reaches AP50 $=0.158$ and Hole-F1 $=0.310$, but with detector proposals AP50 drops to $0.001$ and Hole-F1 to $0.002$ (Table~\ref{tab:app_roipoly_ablation}). The polygon head is functional, and the proposal stage is the main bottleneck.
\end{itemize}
In summary, the AP50 collapse of direct methods is caused by instance discovery rather than hole recovery. All three direct methods were designed for scenes containing a few simple buildings; on dense patches with complex polygons they fail to find and complete the instances in the first place. This is a limitation of how these models represent their outputs rather than of how they are trained, and hyperparameter tuning alone cannot resolve it. These results motivate explicitly supervising interior rings and enforcing ring hierarchy and validity during decoding (Section~\ref{sec:why_fail}).

\subsection{Annotation Protocol and Quality Control}
\label{app:annotation_protocol}

\textbf{Inria Building.}
The Inria Aerial Image Labeling release~\cite{maggiori2017can} provides $180$ tiles of $5000\times5000$ pixels across five cities with binary foreground raster masks but no vector annotation. We take OpenStreetMap~\cite{OpenStreetMap} building footprints as the vector source because they provide topologically clean multi-ring polygons with rich metadata, and correct two known issues before release: (1) $2$--$8$~pixel geo-registration offsets against the imagery, and frequent omission of inner courtyards. To fix the offsets, for every OSM polygon we search a translation within $\pm 16$~pixels that maximises IoU with the overlapping raster connected component, accept the translation when the resulting IoU exceeds $0.5$, and drop polygons without a majority-overlap component. The authors then review the aligned polygons tile by tile and manually correct residual mismatches. To recover missing courtyards, we vectorise the raster mask by tracing connected components and extracting nested rings, then transfer every nested ring that falls inside an aligned OSM exterior as a candidate interior ring; this step produces most of the hole-bearing annotations in Table~\ref{tab:dataset_complexity}. We finally apply automated filters to every polygon. Firstly, self-intersections via \texttt{shapely.buffer(0)}, and reassigning interior rings that are not strictly contained in their exterior via \texttt{shapely.contains}. Tiles are sliced into non-overlapping $512\times512$ patches and split at the tile level ($145/35$ for train/val), stratified by city and hole-bearing fraction, and the split is shared by all baselines.

\textbf{Deventer land cover.}
The Deventer tasks reuse the \emph{Deventer-512} vector annotations of Jiao et al.~\cite{jiao2026acpv}; we select three of their six classes (\emph{Road}, \emph{Vegetation}, \emph{Unvegetated}) because they contain the most multi-ring structure (Table~\ref{tab:dataset_complexity}) and we inherit the released train/val split. We do not re-annotate and only apply the automated filters applied in Inria, and review the annotations by our authors. We refer the reader to~\cite{jiao2026acpv} for the upstream annotation procedure.

\textbf{Limitations.}
Our Inria annotation inherits two error sources. First, OSM itself can be incomplete or temporally mismatched with the imagery, so buildings that appear in only one source are dropped rather than re-annotated; we do not cross-validate against a second independent vector source due to the limited publicly available vector data. Second, the hole-restoration step relies on the Inria raster mask and therefore inherits any omissions or false positives in the official raster label. For the Deventer dataset, we rely on the upstream annotation of~\cite{jiao2026acpv}. Though our authors examine the annotations, some wrong labels may still exist. 

\subsection{Bootstrap Confidence Intervals}
\label{app:bootstrap_ci}

We additionally report image-level bootstrap confidence intervals to estimate the statistical stability of the main benchmark metrics. We first compute the evaluation statistics separately for each image, using the saved predictions and ground truth annotations. For one bootstrap trial on a task with $N$ evaluation images, we sample $N$ image IDs from the original image list. The sampling is with replacement, so the same image ID can be selected more than once and some image IDs may not be selected. We then aggregate the per-image statistics over the sampled image IDs, counting repeated images repeatedly, and compute AP50, E-BIoU, Hole-F1, and Topo-EM for that sampled task. We repeat this process $500$ times and report the macro-average confidence intervals over the four tasks. The results are shown in Table \ref{tab:app_bootstrap_ci}. This analysis measures uncertainty from the finite evaluation set. It is not a multi-seed training study, so it does not capture variation from retraining the same model under different random seeds.

\begin{table}[t]
\centering
\scriptsize
\setlength{\tabcolsep}{3pt}
\caption{Macro-average benchmark scores with 95\% image-level bootstrap confidence intervals over 500 resamples.}
\label{tab:app_bootstrap_ci}
\resizebox{\textwidth}{!}{%
\begin{tabular}{llcccc}
\toprule
Method & Type & AP50 & E-BIoU & Hole-F1 & Topo-EM \\
\midrule
U-Net + Poly. & Seg. & 0.455 [0.437, 0.478] & 0.294 [0.288, 0.299] & 0.357 [0.335, 0.379] & 0.246 [0.235, 0.257] \\
Mask R-CNN + Poly. & Seg. & 0.343 [0.329, 0.361] & 0.260 [0.252, 0.268] & 0.112 [0.100, 0.127] & 0.243 [0.233, 0.254] \\
SAM2 + Poly. & FM & 0.134 [0.128, 0.145] & 0.274 [0.264, 0.285] & 0.011 [0.008, 0.013] & 0.127 [0.121, 0.133] \\
HiSup & Rep. & 0.315 [0.298, 0.338] & 0.256 [0.250, 0.262] & 0.286 [0.263, 0.309] & 0.187 [0.177, 0.198] \\
ACPV-Net & Rep. & 0.323 [0.306, 0.341] & 0.263 [0.256, 0.270] & 0.298 [0.277, 0.318] & 0.189 [0.180, 0.199] \\
FFL & Rep. & 0.207 [0.194, 0.221] & 0.217 [0.207, 0.229] & 0.000 [0.000, 0.000] & 0.096 [0.092, 0.100] \\
GCP & Rep. & 0.296 [0.283, 0.313] & 0.249 [0.242, 0.256] & 0.010 [0.008, 0.012] & 0.025 [0.024, 0.026] \\
HoliTracer & Rep. & 0.043 [0.038, 0.052] & 0.202 [0.194, 0.210] & 0.062 [0.052, 0.072] & 0.053 [0.049, 0.059] \\
Pix2Poly & Direct & 0.033 [0.032, 0.037] & 0.252 [0.205, 0.352] & 0.012 [0.010, 0.014] & 0.028 [0.025, 0.030] \\
PolyWorld & Direct & 0.005 [0.002, 0.005] & 0.146 [0.132, 0.162] & 0.000 [0.000, 0.001] & 0.006 [0.006, 0.007] \\
RoIPoly & Direct & 0.010 [0.008, 0.013] & 0.138 [0.116, 0.162] & 0.082 [0.069, 0.096] & 0.019 [0.017, 0.023] \\
\bottomrule
\end{tabular}}
\end{table}

\subsection{Effect of the Simple--Complex Polygon Imbalance}
\label{app:imbalance}
Table~\ref{tab:dataset_complexity} shows that simple polygons are much more frequent than complex polygons at the instance level. The imbalance is milder at the patch level, which is the actual training unit: $15.2\%$ of the Inria training patches contain at least one complex polygon, and the corresponding proportions on Deventer are $44.0\%$ for Road, $30.7\%$ for Vegetation, and $55.7\%$ for Unvegetated. Models therefore receive interior-ring supervision approximately every second to sixth patch.

The observed failure pattern does not indicate that complex polygons are treated as noise. If models simply ignored these less frequent instances, complex polygons would be missed entirely or localized poorly at the instance level. Instead, many complex instances are successfully matched to their ground truth with IoU $\geq 0.5$, showing that the models recognize the objects and recover their overall extent. Yet $84\%$--$100\%$ of the matched complex polygons still have incorrect ring structures (Figure~\ref{fig:wrong_topo_geometry}), most commonly because one or more interior rings are missing. The models have thus learned the presence and exterior shape of complex polygons but struggle to represent their multi-ring topology.

To measure the effect of the imbalance directly, we retrain U-Net + Poly. on Inria Building while keeping the data, the number of gradient steps, and the random seed fixed, and vary only the sampling probability of hole-bearing patches with weighted random sampling. Table~\ref{tab:app_imbalance} shows that a $16\times$ increase in the sampling weight of hole-bearing patches (from $4.3\%$ to $41.8\%$ of the training stream) raises Hole-F1 from $0.455$ to $0.559$, so oversampling helps detect individual holes. Topo-EM, however, remains nearly unchanged, because it requires the complete ring structure to be correct. The imbalance therefore moderately affects hole detection but is not the main bottleneck for topology correctness. \model{} intentionally preserves the natural class distribution because it reflects how these objects occur in the real world; rebalancing would make the benchmark easier but less faithful.

\begin{table}[t]
\centering
\small
\setlength{\tabcolsep}{4pt}
\caption{Effect of hole-bearing patch sampling on U-Net + Poly. (Inria Building). Only the sampling probability of patches containing at least one complex polygon is changed; data, gradient steps, and seed are fixed. The natural distribution corresponds to the main benchmark setting.}
\label{tab:app_imbalance}
\begin{tabular}{lcccc}
\toprule
Hole-bearing patches in the training stream & AP50 & E-BIoU & Hole-F1 & Topo-EM \\
\midrule
Down-sampled ($4.3\%$) & 0.678 & 0.304 & 0.455 & 0.529 \\
\textbf{Natural ($15.2\%$)} & \textbf{0.670} & \textbf{0.308} & \textbf{0.514} & \textbf{0.531} \\
Up-sampled ($41.8\%$) & 0.679 & 0.304 & 0.559 & 0.533 \\
\bottomrule
\end{tabular}
\end{table}


\end{document}